\documentclass[preprint,12pt]{elsarticle}

\usepackage{amssymb}
\usepackage{lmodern}
\usepackage{graphicx} %
\usepackage{booktabs}
\usepackage{longtable}
\usepackage{makecell}
\usepackage{comment}
\usepackage{amsmath,amsfonts,bm}
\usepackage[edges]{forest}
\usepackage{changepage}
\usepackage{geometry}
\usepackage{indentfirst}
\usepackage[breaklinks=true]{hyperref}
\usepackage{breakcites}
\newcommand\remi{}
\newcommand\remibis{}
\newcommand\gianni{}

\journal{Information Fusion}

\begin{document}

\begin{frontmatter}

\title{Explainability for Vision Foundation Models: A Survey} %

\author[label1]{Rémi Kazmierczak \corref{cor1}} %
\author[label1]{Eloïse Berthier}
\author[label1]{Goran Frehse}
\author[label1]{Gianni Franchi}

\cortext[cor1]{remi.kazmierczak@ensta-paris.fr}

\affiliation[label1]{organization={U2IS, ENSTA, Institut Polytechnique de Paris}}

\begin{abstract}
\remibis{As artificial intelligence systems become increasingly integrated into daily life, the field of explainability has gained significant attention. This trend is particularly driven by the complexity of modern AI models and their decision-making processes. The advent of foundation models, characterized by their extensive generalization capabilities and emergent uses, has further complicated this landscape. Foundation models occupy an ambiguous position in the explainability domain: their complexity makes them inherently challenging to interpret, yet they are increasingly leveraged as tools to construct explainable models. In this survey, we explore the intersection of foundation models and eXplainable AI (XAI) in the vision domain. We begin by compiling a comprehensive corpus of papers that bridge these fields. Next, we categorize these works based on their architectural characteristics. We then discuss the challenges faced by current research in integrating XAI within foundation models. Furthermore, we review common evaluation methodologies for these combined approaches. Finally, we present key observations and insights from our survey, offering directions for future research in this rapidly evolving field.}
\end{abstract}

\begin{highlights}
\item Compilation and organization of a comprehensive corpus of articles at the intersection of XAI and foundation models in vision.
\item Identification and presentation of trends in XAI evaluation methodologies associated with these approaches.
\item In-depth analysis of emerging implications stemming from the integration of foundation models into XAI.
\item Highlighting key challenges and future directions for advancing the domain.
\end{highlights}


\end{frontmatter}

\section{Introduction}

\begin{figure}[ht]
    \centering
    \includegraphics[width=0.8\textwidth]{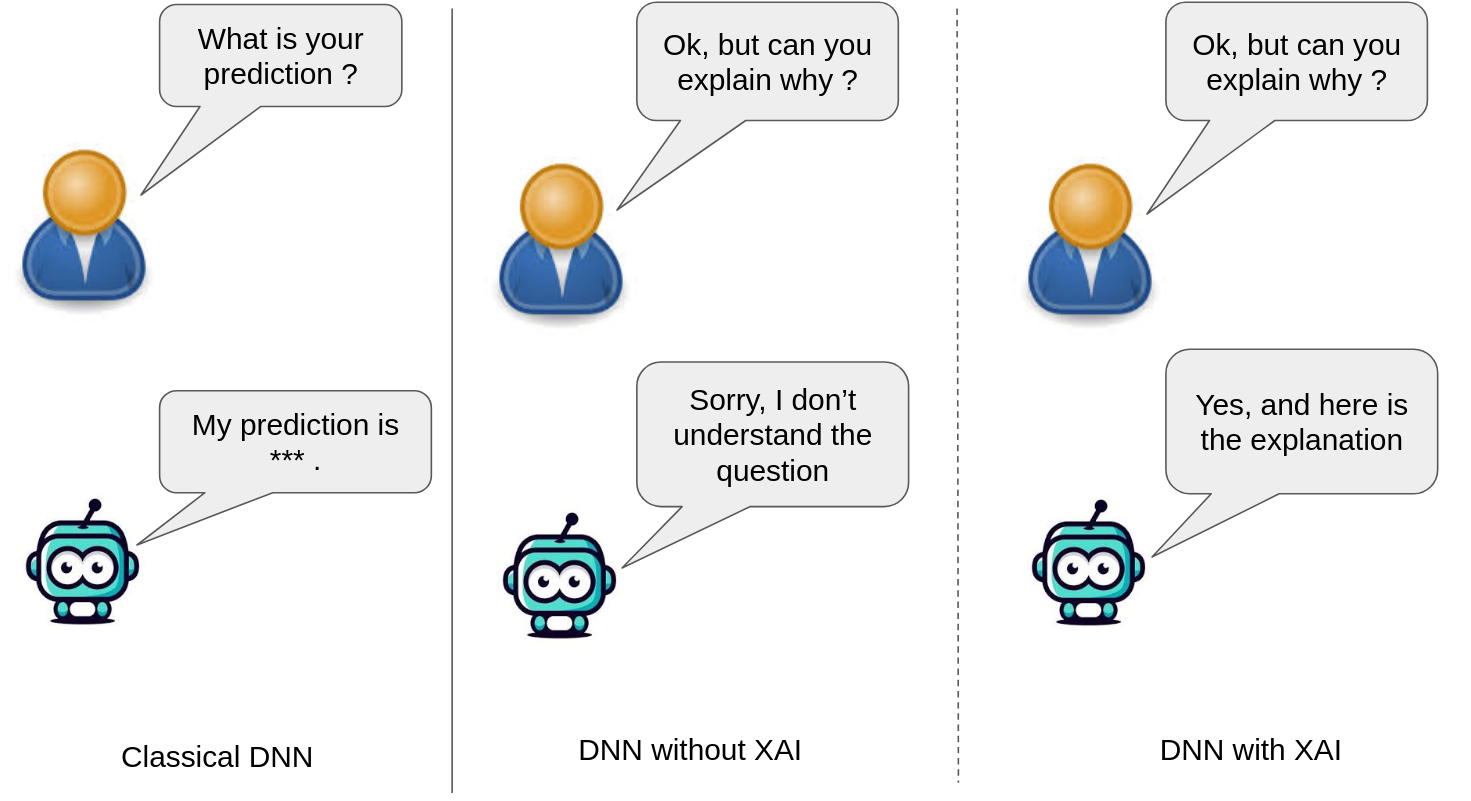}
    \caption{\remi{\textbf{Global goal of XAI.} While a non explainable method only makes inference, an explainable model produces details about the reasons of its decisions, to make its functioning clear or easy to understand}}
    \label{fig:princip_XAI}
\end{figure}

Deep Neural Networks (DNNs), i.e., networks with a large number of trainable parameters, have had a significant impact on computer vision in recent years \cite{lecun2015deep}. 
They have achieved state-of-the-art performance in various tasks such as semantic segmentation \cite{wang2023internimage}, classification \cite{wortsman2022model}, and image generation \cite{sauer2022stylegan}. 
However, the depth and complexity of DNNs also lead to a lack of transparency~\cite{castelvecchi2016can} %
in decision-making and in the interpretability of predictions~\cite{arrieta2020explainable}. %
There is an increasing demand for %
\remibis{transparent} DNN models in high-stakes environments where both performance and interpretability are crucial \cite{preece2018stakeholders}. A wide range of approaches that add transparency and interpretability is broadly referred to as eXplainable Artificial Intelligence (XAI) \cite{rawal2021recent} \remi{(see Figure \ref{fig:princip_XAI})}.

\remi{
XAI methods provide a bridge between an automated system and human users, whose perceptions and interpretations are inherently subjective. 
An explanation that satisfies one user may not necessarily satisfy another \cite{yang2022psychological},
so to be effective XAI methods should ensure consistency in the interpretations across different users \cite{wang2019designing}.
XAI has garnered increasing interest, particularly in fields where ethical concerns are paramount, such as medical diagnosis \cite{van2022explainable} and autonomous driving \cite{atakishiyev2021explainable}, since opaque models may conceal functionalities that contradict moral principles. For instance, gender-biased outcomes have been observed in \cite{hendricks2018women}. 

Several properties have been identified in the literature as essential for XAI
\cite{nauta2023anecdotal,arrieta2020explainable}, such as trustworthiness, complexity, robustness, generalizability, and \remibis{objectiveness}. We explore this issue further in Section \ref{Axioms_XAI}.

\remi{A noticeable trend in deep learning is the use of models that are larger and larger (see Figure \ref{fig:evol_DNN}). The trend began in computer vision with LeNet (60,000 parameters) in 1998, then InceptionV3 (6.23M parameters) in 2014, and then Resnet (42.70M parameters) in 2016. Then,  the field of natural language processing followed with Transformers (65M parameters) in 2017, then BERT (340M parameters) in 2018, then GPT-2 (1.5T parameters) in 2019, and then QWEN (72B parameters) in 2023.}
\remibis{
The success of %
these 
``large language models''
has sparked interest in applying the benefits of high parameter counts and extensive training data to other domains, such as visual question answering \cite{liu2024visual} and object detection \cite{liu2025grounding}. This %
has led to the broader classification of such architectures under the global term ``foundation models.''}
Foundation models are in an ambiguous position in the XAI field. 
\remi{%
On the one hand, the complexity of foundation models makes them particularly difficult to explain. On the other hand, they become increasingly used in the literature as tools to build explainable models.}

\begin{figure}
    \centering
    \includegraphics[width=\textwidth]{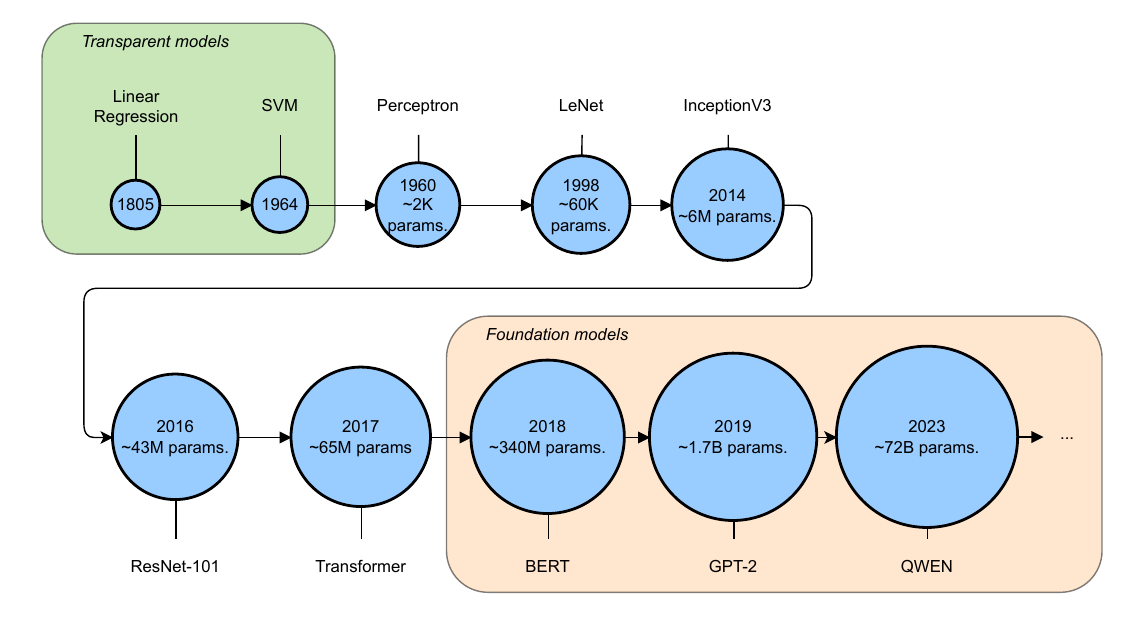}
    \caption{\remi{\textbf{Chronology of the order of magnitude of the number of parameters of learning methods.} While early methods were interpretable and lightweight, subsequent developments have led to an increase in complexity that has culminated in foundation models, which are mainly characterized by their size.}}
    \label{fig:evol_DNN}
\end{figure}

This survey provides a panorama of explainability techniques within the field of foundation models in computer vision, and more particularly \emph{pretrained foundation models}(PFMs). 
It is structured as follows.
\remibis{Section~\ref{background} provides background on foundation models and XAI methods, takes stock of existing surveys, and proposes a taxonomy for XAI methods. Section~\ref{PFM_for_XAI} defines the identified classes of XAI methods and describes their background, their use of PFMs, their applications, and their evaluation. In Section~\ref{XAI_evaluation}, we discuss different methods used to evaluate the quality of produced explanations. Some observations from our survey are presented in Section \ref{Observations}. The different challenges faced by XAI methods are described in Section \ref{sec:limits_and_challenges}, including a description of problems that remain open. Finally, Section~\ref{Conlusion} presents our conclusions and potential avenues for further research.}

\section{Background} \label{background}

In this section, we delineate the scope of our investigation through the definition of the main terminologies. Subsection \ref{background_PFMs} presents the range of vision foundation models, offering a comprehensive understanding of their fundamental aspects. Subsection \ref{background_XAI} refers to the contextual background associated with XAI and its interrelation with interpretability. \remibis{Subsection \ref{exist_survey} points out similar surveys and how the present one differs from them.} Lastly, Subsection \ref{background_formalism} presents a spectrum of distinct strategies aimed at amalgamating foundation models with XAI methodologies.

\subsection{Foundation models} \label{background_PFMs}

\remi{According to \cite{schneider2024foundation}, a foundation model is defined as ``any model trained on broad data that can be adapted to a wide range of downstream tasks.'' The significance of foundation models lies in their demonstrated ability to generalize across diverse tasks, leading to their emergent utilization in various applications. While there is widespread consensus to label deep learning models like GPT \cite{brown2020language} or CLIP \cite{radford2021learning} as foundation models, a significant debate persists regarding the delineation between foundation models and other DNNs models \cite{zhou2023comprehensive}. The definition pivots on the notion of extensive datasets and a high volume of data, a criterion that remains subjective and open to interpretation.}

While the concept of foundation model is relatively recent, the use of large pretrained models to enhance performance predates their emergence. Ancestral to foundation models, the widely employed 
\remi{feature representation backbones}, such as VGG \cite{simonyan2014very}, ResNet \cite{he2016deep}, and ViT \cite{dosovitskiy2020image} pretrained on ImageNet \cite{deng2009imagenet}, \remi{paved the way of modern computer vision with DNNs}. \gianni{These pretrained feature representation backbones were used as an initialization for other vision tasks \cite{zhao2018psanet}.} 
\remi{Indeed, it has been observed that incorporating such techniques facilitates faster convergence, particularly in scenarios with limited training data \cite{he2019rethinking}. Another emphasis of foundation models is the prevalent use of self-supervised techniques to benefit from the widest corpus of data.}

The initial models recognized as foundation models emerged from the domain of large language models \remi{based on transformers \cite{vaswani2017attention},} such as GPT-2 \cite{radford2019language} and BERT \cite{devlin2018bert}. Their designation as such stemmed from the fact that these large language models exhibit remarkable generalization performances, coupled with their exploitation of extensive text datasets. This success has subsequently led to various emerging applications \cite{li2020contextualized, ravichander2020systematicity}.

The success observed in large language models \remi{encouraged} a shift towards scaling models in other domains, notably in computer vision \cite{liu2024visual,li2023blip}, where substantial volumes of data are readily accessible. Moreover, the widespread adoption of transformers in vision tasks aligns with the structural foundation laid by language models. Consequently, this evolution has led to a new wave of models. \remibis{Notably,} language/vision models, capable of processing textual and visual inputs and projecting them into a shared embedding space. A prominent example of this category is CLIP \cite{radford2021learning}, known for its ability to represent \remi{text and language} modalities within a unified framework. \remi{Also,}  text-conditioned diffusion models, capable of generating high-quality images across a diverse spectrum of tasks \remi{like stable diffusion \cite{rombach2021high}}. These models demonstrate proficiency in creating images through conditioning on extensive textual information, marking a significant advancement within the domain.

After the success achieved by leveraging diverse modalities and driven by the aspiration to craft increasingly versatile agents, a new wave of foundation models has emerged. These models endeavor to incorporate an expanding array of modalities, exemplified by innovations like IMAGEBIND \cite{girdhar2023imagebind} and GATO \cite{reed2022generalist}. These advanced foundation models not only process text and images but also integrate additional elements such as sound or action data, enriching their capacity to comprehend varied inputs across multiple \remi{modalities}. Furthermore, the scaling-up trend has extended beyond conventional data sources, embracing more challenging and intricate datasets. For instance, there has been a notable trend towards scaling up models to handle more \remi{diverse tasks. %
\remibis{A widely explored task in this context is visual question answering, which leverages the capabilities of large language models by incorporating visual tokens. For instance, LLaVa \cite{liu2024visual} extends the LLaMa architecture \cite{touvron2023llama} to handle visual data, BLIP \cite{li2023blip} adapts BERT \cite{devlin2018bert} for image-based tasks. Another rapidly growing area is zero-shot object detection and segmentation, exemplified by recent models such as SAM \cite{zou2024segment}, SAM2 \cite{ravi2024sam}, and Grounding DINO \cite{liu2025grounding}. \remibis{Finally, the trend to scale up models has progressed, as shown by the examples of GPT-4 \cite{achiam2023gpt} or Pixtral \cite{agrawal2024pixtral}}}

These models are commonly used in a pretrained modality, where their weights remain fixed, hence the term Pretrained Foundation Models (PFMs). Subsequently, there are two primary approaches to their utilization. First, these models can be fine-tuned by training a lightweight probe on top of them \remi{(or on top of their intermediate feature embeddings). Another technique is to use a low rank adaptator to fine tune the PFM \cite{hu2021lora}.} Alternatively, they can be employed end-to-end to execute specific tasks, facilitated by conditioning techniques such as prompt engineering. 
For a more visual representation, a comprehensive non exhaustive summary of these models is depicted in Table~\ref{table:PFMs}.

We delineate the scope of our investigation from the range of existing PFMs as follows:
\begin{itemize}
\item we include methods using vision PFMs: these models are characterized by their \remi{handling of the vision modality}, including images or videos, either as input or output modalities;
\item we exclude backbones pretrained \remi{exclusively} on ImageNet due to the ambiguity surrounding their classification as PFMs.
\end{itemize}

\begin{table}[t]
\fontsize{8.5pt}{8.5pt}\selectfont
\caption{\textbf{Overview of Different PFMs.} \textbf{Input} refers to the modalities of the input, \textbf{Output} refers to the modality of the output, and \textbf{Type} refers to the categorization of PFMs according to the formalism of \cite{speith2022review}.}
\label{table:PFMs}
\centering
\begin{tabular}{llll}
\toprule
\textbf{Method} & \textbf{Input} & \textbf{Output} & \textbf{Type} \\
\midrule
Bert \cite{devlin2018bert} & Text & Text & Generative \\
BLIP-2 \cite{li2023blip} & Text, Image & Text & Hybrid \\
CLIP \cite{radford2021learning} & Text, Image & Similarity score & Contrastive \\
CoCa \cite{yu2022coca} & Text, Image & Text & Hybrid \\
Dalle-2 \cite{ramesh2022hierarchical} & Text & Image & Generative \\
Dalle-3 \cite{betker2023improving} & Text & Image & Generative \\
Flamingo \cite{alayrac2022flamingo} & Text, Image & Text & Conversational \\
GiT \cite{wang2022git} & Image & Text & Adaptation \\
Glide \cite{nichol2021glide} & Text, Image & Image & Generative \\
GPT4 \cite{achiam2023gpt} & Text, Image & Text & Conversational \\
GroundingDINO & \cite{liu2025grounding} Image & Bounding Boxes & \remibis{Contrastive} \\
LLaVA \cite{liu2024visual} & Text, Image & Text & Conversational \\
LXMERT \cite{tan2019lxmert} & Text, Image & Text & Conversational \\
MiniGPT4 \cite{zhu2023minigpt} & Text, Image & Text & Conversational \\
Mplug \cite{li2022mplug} & Text, Image & Text & Generative \\
OFA \cite{wang2022ofa} & Text, Image & Text & Conversational \\
Segment Anything \cite{kirillov2023segment} & Image & Instance segmentation & Foundational \\
Stable Diffusion \cite{rombach2021high} & Text & Image & Generative \\
STAIR \cite{chen2023stair} & Text, Image & Text & Hybrid \\
VisualBERT \cite{li2019visualbert} & Text, Image & Text & Conversational \\
\bottomrule
\end{tabular}
\end{table}

\subsection{Explainable AI (XAI)} \label{background_XAI}

\begin{figure}
    \centering
    \includegraphics[width=0.5\textwidth]{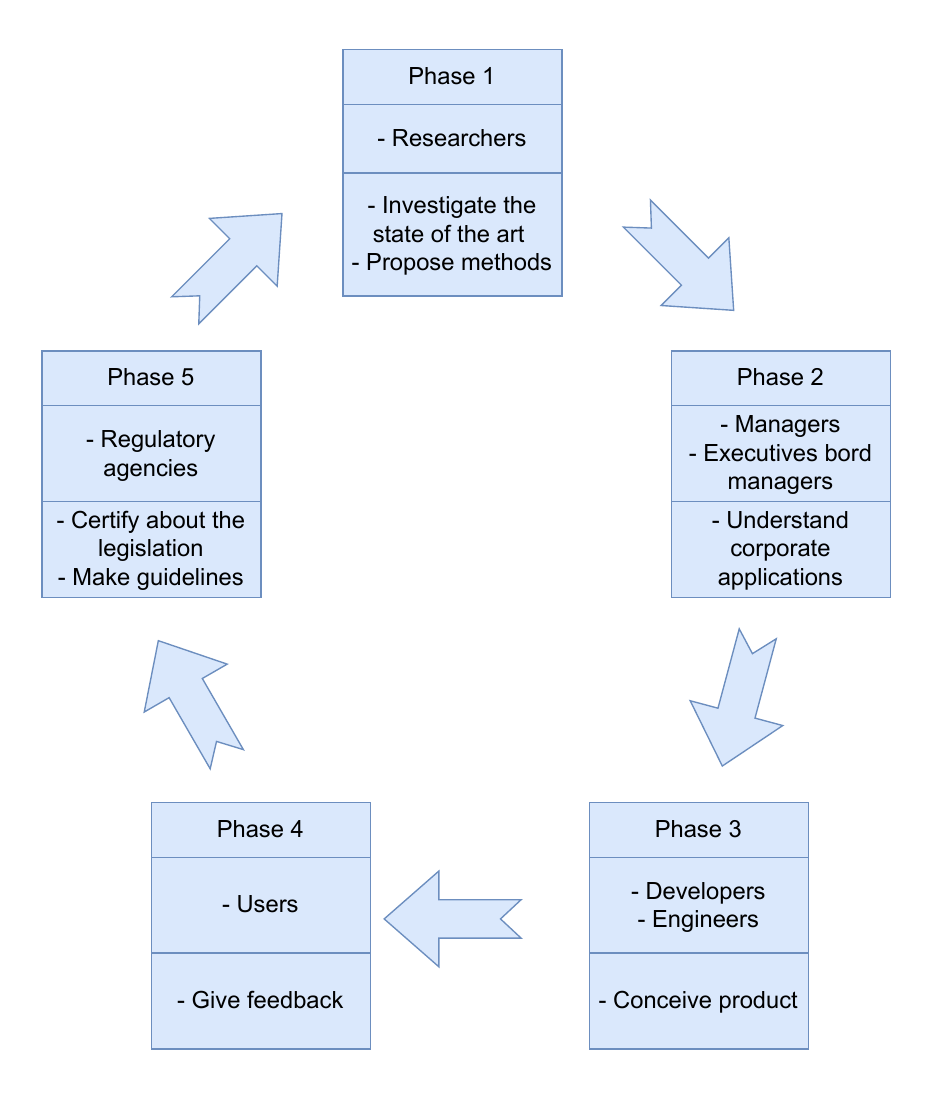}
    \caption{\remi{\textbf{Chain representing the acceptance of AI in society.} Each box presents the involved audience (middle) and the description of the step (bottom).}}
    \label{fig:society_XAI}
\end{figure}

According to \cite{arrieta2020explainable}, we can define an explainable model as a computational model that is designed to provide specific details or reasons regarding its functioning, to ensure clarity and ease of understanding. In broader terms, an explanation denotes the information or output that an explainable model delivers to elucidate its operational processes.

\remi{In the literature, as} notably highlighted in \cite{zhao2024explainability, poeta2023concept}, there exists a nebulous distinction between the terminologies ``interpretability'' and ``explainability''. In some instances, these terms are used interchangeably, further complicating their differentiation. To ensure coherence and eliminate ambiguity, 
\remi{we choose to use the terms explainable and interpretable synonymously.}

From a historical perspective, the primary explainability methods for early AI algorithms involved employing transparent models. Such models are characterized by their simplicity, which allows their decision process to serve as an explanation in itself. These models are easily interpretable due to their straightforward nature and clear features. Examples of such models include linear regression \cite{galton1886regression}, logistic regression \cite{mccullagh2019generalized}, and decision trees \cite{quinlan1986induction}. For instance, an explanation generated by a decision tree consists of a series of logical assertions that lead to the selection of a specific leaf node, \remi{narrowing the gap with neural symbolic AI \cite{garcez2022neural}}. However, as noted by previous research, the explainability potential of these methods is contingent upon the complexity of their construction: if the number of parameters becomes too large, transparency is compromised.

In particular, the pursuit of performance, as evidenced by the competition to achieve the highest ImageNet top-1 accuracy, has led to the development of models with increasingly large numbers of parameters. Consequently, state-of-the-art methods have gained a reputation for being opaque, as their inner workings are often incomprehensible to humans. In response to this challenge, additional techniques known as post-hoc methods have emerged \cite{lime,selvaraju2017grad,petsiuk2018rise}. These methods are applied to the model after the training process to provide explanations. Commonly used post-hoc methods \remi{include different approaches such as visualization techniques \cite{selvaraju2017grad}}, which highlight influential \remi{parts} in an image that contribute most to the model's decision-making. Sensitivity analysis~\cite{cortez2011opening} represents another approach, based on the analysis of the variations of the model's predictions when the input data change. Local explanation techniques, such as LIME \cite{lime}, aim to explain the model's predictions by creating a local, simplified model around a point of interest, \remi{which is transparent}. Finally, feature relevance techniques, such as SHAP \cite{lundberg2017unified}, estimate the impact of each feature on the model's decision.

\remi{In opposition to post-hoc ones, ante-hoc methods produce explanations by design \cite{speith2022review}.} With the growing availability of models capable of performing auxiliary tasks and architectures structured as a sequence of subtasks, there has been a shift towards what \cite{akhtar2023survey} describes as "inherently explainable models." These models, while not inherently transparent, incorporate interpretable components that facilitate human understanding. \remi{It is noteworthy that inherently explainable models are not transparent in nature. Instead, they achieve interpretability through the incorporation of interpretable components, in a way that makes their functioning understandable by humans.} A typical example of such models is those based on Chains of Thought reasoning (see Section \ref{CoT_PFM_for_XAI}). These models, while being complex and opaque, are considered interpretable because they provide textual hints in addition to their output, helping to understand their functioning. \remi{Another example is Concept Bottleneck Models (Section \ref{sec:cbms}), which, while not inherently transparent, are designed to describe inputs using semantically interpretable concepts. Similarly, Prototypical Networks (Section \ref{sec:prototypes}) learn semantically meaningful prototypes during training, providing an additional layer of interpretability. These families of models allow for the integration of various interpretability tools. For instance, logical reasoning can be incorporated into Concept Bottleneck Models to process and analyze concepts or used to establish relationships between components in chain-of-thought-based models.}

We delineate the scope of our study from the spectrum of available XAI methods as follows:
\begin{itemize}
\item Our focus is solely on XAI methods used in conjunction with vision PFMs. This encompasses the corpus of available PFMs, as constrained by the scope defined in Section \ref{background_PFMs}.
\item We examine papers \remi{focusing on} either ``interpretable'' or ``explainable'' artificial intelligence, without differentiation.
\end{itemize}
 It is important to note that transparent methods are absent from our study by design, as the presence of PFMs inherently leads to opaque models. Then, our study is categorized into (1) PFMs to facilitate XAI methods, whether as post-hoc methods or inherently explainable models \remibis{(see Section \ref{PFM_for_XAI})}, and \remi{(2) papers tackling issues and challenges about explaining PFMs (see Section \ref{sec:limits_and_challenges}).} For further insights about transparent models, we redirect the reader to the survey of \cite{arrieta2020explainable}.

\subsection{Existing surveys} \label{exist_survey}

The burgeoning need for feedback on AI models has resulted in a substantial surge in publications in the domain of XAI. This escalation is highlighted by the emergence of meta-surveys and comprehensive analyses, reflecting the growing landscape of research in this area \cite{chatzimparmpas2020survey,saeed2023explainable}. Numerous prior studies have delved into subjects closely aligned with our investigations. For instance, \cite{joshi2021review} focuses specifically on elucidating explainability within multimodal contexts, while \cite{zhao2024explainability} centers on the explainability of large language models. Additionally, \cite{speith2022review} addresses related \remi{taxonomies}, underscoring the breadth and depth of prior research relevant to our study. \remi{Compared to existing works, our survey emphasizes on the recent use of vision PFMs in XAI.}

\subsection{\remi{Corpus}} \label{background_formalism}

\paragraph{Methodology} \label{methodology}

\remi{To gain a comprehensive understanding of how PFMs are utilized in XAI methods, we began by assembling a corpus of relevant papers. This corpus comprises 122 studies, including 76 on inherently explainable models (Section \ref{sec:inherently_interpretable_methods}), 20 on post-hoc methods (Section \ref{sec:post_hoc_methods}), and 26 papers addressing enhancing the explainability of PFMs (Section \ref{sec:limits_and_challenges}). All the selected papers were published until \remibis{January 2025}.}

\paragraph{Taxonomy} \label{taxonomy}

In prior studies, the need for adaptable organizational frameworks in the large field of XAI has led to the introduction of various taxonomies. As presented in \cite{speith2022review}, which extensively examines diverse taxonomies across surveys, prevalent approaches encompass stages, types of results, functioning approaches, output formats of explanations, and scope.

\remi{Drawing upon the definitions provided in \cite{speith2022review}, we separate each of the methods of our corpus. The resulting taxonomy is presented in Figure \ref{taxonomy_general} and \ref{taxonomy_chall}, as well as Tables \ref{xai-methodsPFMforXAI} and \ref{xai-methods-posthoc}.}
A detailed characterization and discussion of each of the identified categories is provided in the next section.

\begin{table}[htbp]
\fontsize{8.5pt}{8.5pt}\selectfont
\caption{\textbf{Overview of inherently explainable models.} \textbf{Scope} refers to \remi{whether the method offers insights at the sample level (local) or across the entire dataset (global).} \textbf{Format} refers to the modality of the explanation. \textbf{Functioning} refers to the separation described in the taxonomy of \cite{speith2022review}. \textbf{Result} refers to the type of explanation.}
\label{xai-methodsPFMforXAI}
\centering
\begin{tabular}{@{}lllll@{}}
\toprule
\textbf{Method} & \textbf{Scope} & \textbf{Format} & \textbf{Functioning} & \textbf{Result} \\
\midrule
Adaptative CBM \cite{choi2024adaptive} & Local & Textual & Leveraging structure & Feature importance \\
ARTxAI \cite{fumanal2023artxai} & Local & Textual & Leveraging structure & Feature importance \\
Beyond Accuracy \cite{libeyond} & Local & Textual & Leveraging structure & Feature importance \\
BBA \cite{zhao2024bba} & Local & Textual & Leveraging structure & Feature importance \\
CBM with filtering \cite{kim2023concept} & Local & Numerical, Textual & Leveraging structure & Feature importance \\
CC:DAE \cite{achille2024interpretable} & Local & Textual & Leveraging structure & Feature importance \\
CEIR \cite{cui2023ceir} & Local & Textual & Leveraging structure & Feature importance \\
Chat GPT XAI \cite{liu2023chatgpt} & Local & Textual, Visual & Leveraging structure & Feature importance \\
ChatGPT CBM \cite{ren2024chatgpt} & Local & Numerical, Textual & Leveraging structure & Feature importance \\
ChartThinker \cite{liu2024chartthinker} & Local & Textual & Leveraging structure & Feature importance \\
Classification with descriptors \cite{menon2022visual} & Local & Textual & Leveraging structure & Feature importance \\
CLIP-QDA \cite{kazmierczakclip} & Local, Global & Num., Textual, Visual & Leveraging structure & Feature importance \\
CoBRa \cite{ZhangLearning} & Local & Rules & Leveraging structure & Feature importance \\
Concept Gridlock & Local & Textual & Leveraging structure & Feature importance \\
ComFe \cite{mannix2024scalable} & Local & Visual & Leveraging structure & Feature importance \\
CoT Prompt \cite{ge2023chain} & Local & Textual & Leveraging structure & Feature importance \\
CoT diffusion \cite{harvey2023visual} & Local & Textual & Leveraging structure & Feature importance \\
CoTBLIP \cite{chen2023measuring} & Local & Textual & Leveraging structure & Feature importance \\
CPSeg \cite{li2024cpseg} & Local & Textual, Visual, Rules & Leveraging structure & Feature importance \\
DCLUB \cite{fu2023dynamic} & Local & Textual & Leveraging structure & Feature importance \\
DDCoT \cite{zheng2023ddcot} & Local & Textual & Leveraging structure & Feature importance \\
Decap \cite{wang2023decap} & Local & Rules & Leveraging structure & Feature importance \\
DME-Driver \cite{han2024dme} & Local & Textual & Leveraging structure & Feature importance \\
DoT \cite{grover2024navigating} & Local & Textual & Leveraging structure & Feature importance \\
Dolphins \cite{ma2023dolphins} & Local & Textual & Leveraging structure & Feature importance \\
DriveGPT4 \cite{xu2023drivegpt4} & Local & Textual & Leveraging structure & Feature importance \\
DVP \cite{han2024image} & Local & Textual, Visual & Leveraging structure & Feature importance \\
ECEnet \cite{zhang2023ecenet} & Local & Textual & Leveraging structure & Feature importance \\
Explainable meme classification \cite{thakur2022multimodal} & Local & Visual & Examples & Examples \\
Explicit CoT \cite{uehara2024advancing} & Local & Textual & Leveraging structure & Feature importance \\
ExTraCT \cite{yow2024extract} & Local & Textual & Leveraging structure & Feature importance \\
GenSAM \cite{hu2024relax} & Local & Visual, Textual & Leveraging structure & Feature importance \\
GPT4 street crossing \cite{hwang2024safe} & Local & Textual & Leveraging structure & Feature importance \\
Hierarchical CBM \cite{panousis2023hierarchical} & Local & Numerical, Textual & Leveraging structure & Feature importance \\
II-MMR \cite{kil2024ii} & Local & Textual & Leveraging structure & Feature importance \\
Interpretable Cancer screening \cite{ando2023interpretable} & Local & Numerical & Leveraging structure & Feature importance \\
Interpretable VQA \cite{fu2023interpretable} & Local & Rules & Leveraging structure & Feature importance \\
Judge MLLM \cite{lin2024towards} & Local & Textual & Leveraging structure & Feature importance \\
KAM-CoT \cite{mondal2024kam} & Local & Textual & Leveraging structure & Feature importance \\
Label free CBM \cite{oikarinen2023label} & Local, Global & Numerical, Textual & Leveraging structure & Feature importance \\
LaBo \cite{yang2023language} & Local & Textual & Leveraging structure & Feature importance \\
LACA \cite{chen2023reason} & Local & Textual & Architecture Modif. & Feature importance \\
Latent SD \cite{li2023self} & Local & Visual & Architecture Modif. & Feature importance \\
Learning Concise \cite{yan2023learning} & Local, Global & Numerical, Textual & Leveraging structure & Feature importance \\
LEXIS \cite{kassab2023language} & Local & Textual & Leveraging structure & Feature importance \\
LLM Grounded diffusion \cite{lian2023llm} & Local & Textual & Architecture Modif. & Feature importance \\
LLM-Mutate \cite{chiquier2024evolving} & Local & Visual, Textual & Leveraging structure & Feature importance \\
MCLE \cite{lai2024towards} & Local & Textual & Leveraging structure & Feature importance \\
Med-MICN \cite{hutowards} & Rules, Local & Textual & Leveraging structure & Feature importance \\
MMCBM \cite{wu2024concept} & Local, Global & Textual, Visual & Leveraging structure & Feature importance \\
Multimodal VQA \cite{zhu2023multimodal} & Local & Textual & Leveraging structure & Feature importance \\
Q-SENN \cite{norrenbrock2024q} & Local, Global & Visual+Textual & Examples & Feature importance \\
R-VLM \cite{xu2023retrieval} & Local & Visual, Rules & Leveraging structure & Feature importance \\
Reason2Drive \cite{nie2023reason2drive} & Local & Textual & Leveraging structure & Feature importance \\
Robust CBM \cite{yan2023robust} & Local & Num., Textual, Visual & Leveraging structure & Feature importance \\
Skin lesion CBM \cite{patricio2023towards} & Local & Numerical, Textual & Leveraging structure & Feature importance \\
SLOG \cite{banerjee2024learning} & Local & Textual & Leveraging structure & Feature importance \\
SNIFFER \cite{qi2024sniffer} & Local & Textual & Leveraging structure & Feature importance \\
Socratic Reasoning \cite{qi2023art} & Local & Textual, Rules & Leveraging structure & Feature importance \\
Sparse CBM \cite{panousis2023sparse} & Local & Numerical, Textual & Leveraging structure & Feature importance \\
SPANet \cite{wan2024interpretable} & Local & Textual, Visual & Leveraging structure & Feature importance \\
SpatialVLM \cite{chen2024spatialvlm} & Local & Textual, Visual & Leveraging structure & Feature importance \\
SpLiCE \cite{bhalla2024interpreting} & Local & Textual & Leveraging structure & Feature importance \\
STAIR \cite{chen2023stair} & Local & Numerical, Textual & Leveraging structure & Feature importance \\
Stochastic CBM \cite{vandenhirtzstochastic} & Local & Textual & Leveraging structure & Feature importance \\
sViT \cite{kim2024vision} & Local & Visual & Leveraging structure & Feature importance \\
Text-To-Concept \cite{moayeri2023text} & Local & Visual+Textual & Leveraging structure & Feature importance \\
ToA \cite{liang2024toa} & Local & Rules & Leveraging structure & Feature importance \\
VALE \cite{natarajan2024vale} & Local & Textual & Leveraging structure & Feature importance \\
VAMOS \cite{wang2023vamos} & Local & Textual & Local Perturbations & Examples \\
VISE \cite{meng2024few} & Local & Textual, Visual & Leveraging structure & Feature importance \\
Visual CoT \cite{rose2023visual} & Local & Textual, Visual, Rules & Local Perturbations & Examples \\
Visual CoT \cite{shao2024visual} & Local & Visual, Textual & Leveraging structure & Feature importance \\
VLG-CBM \cite{srivastavavlg}  & Local & Textual & Leveraging structure & Feature importance \\
VoroNav \cite{wu2024voronav} & Local & Textual & Leveraging structure & Feature importance \\
XCoOp \cite{bie2024xcoop} & Local & Textual & Leveraging structure & Feature importance \\
\bottomrule
\end{tabular}
\end{table}

\begin{table}[htbp]
\fontsize{8.5pt}{8.5pt}\selectfont
\caption{\textbf{Overview of Post-hoc XAI Methods.} \textbf{Scope} refers to whether the method provides sample-wise insights (local) or dataset-wise insights (Global). \textbf{Output format} refers to the modality of the explanation. \textbf{Functioning} refers to the separation described in the taxonomy of \ref{taxonomy}. \textbf{Result} refers to the type of explanation.}
\label{xai-methods-posthoc}
\centering
\vspace{1em}
\begin{tabular}{lllll}
\toprule
\textbf{Method} & \textbf{Scope} & \textbf{Output Format} & \textbf{Functioning} & \textbf{Result} \\
\midrule
Assessing bias vis \cite{arias2023assessing} & Global & Numerical & Meta Explanation & Feature importance \\
Beyond CLIP \cite{balasubramanian2024decomposing} & Global & Textual, Visual  & Meta Explanation & Feature importance \\
CLIP-Dissect \cite{oikarinen2022clip} & Global & Textual, Visual & Meta Explanation & Feature importance \\
Concept Sliders \cite{gandikota2023concept} & Local & Visual & Leveraging structure & Examples \\
Counterfact latent diffusion \cite{farid2023latent} & Local & Visual & Examples & Examples \\
Decoupling Pixel Flipping \cite{blucher2024decoupling} & Local & Visual & Perturbation & Feature importance \\
Diffusion visual counterfact \cite{augustin2022diffusion} & Local & Visual & Examples & Examples \\
Distilling model failures \cite{jain2022distilling} & Global & Textual, Visual & Meta Explanation & Examples \\
Explain Any Concept \cite{sun2024explain} & Local & Visual, Numerical & Architecture Modification & Surrogate models \\
FALCON \cite{kalibhat2023identifying} & Global & Textual, Visual & Meta Explanation & Examples \\
Grounding counterfactuals \cite{kim2023grounding} & Local & Numerical, Textual & Examples & Feature importance \\
Hard Prompts Easy \cite{wen2024hard} & Local & Textual, Visual & Examples & Examples \\
RePrompt \cite{wang2023reprompt} & Local & Textual, Visual & Local Perturbations & Feature importance \\
SSD-LLM \cite{luo2025llm} & Global & Textual, Visual & Meta Explanation & Examples \\
TeDeSC \cite{liu2024organizing} & Global & Textual, Visual & Meta Explanation & Examples \\
TIFA \cite{hu2023tifa} & Local & Rules & Meta Explanation & Feature importance \\
VIEScore \cite{ku2023viescore} & Local & Textual & Meta Explanation & Feature importance \\
Zero-shot Model Diagnosis \cite{luo2023zero} & Local & Visual, Textual & Examples & Feature importance \\

\bottomrule
\end{tabular}
\end{table}

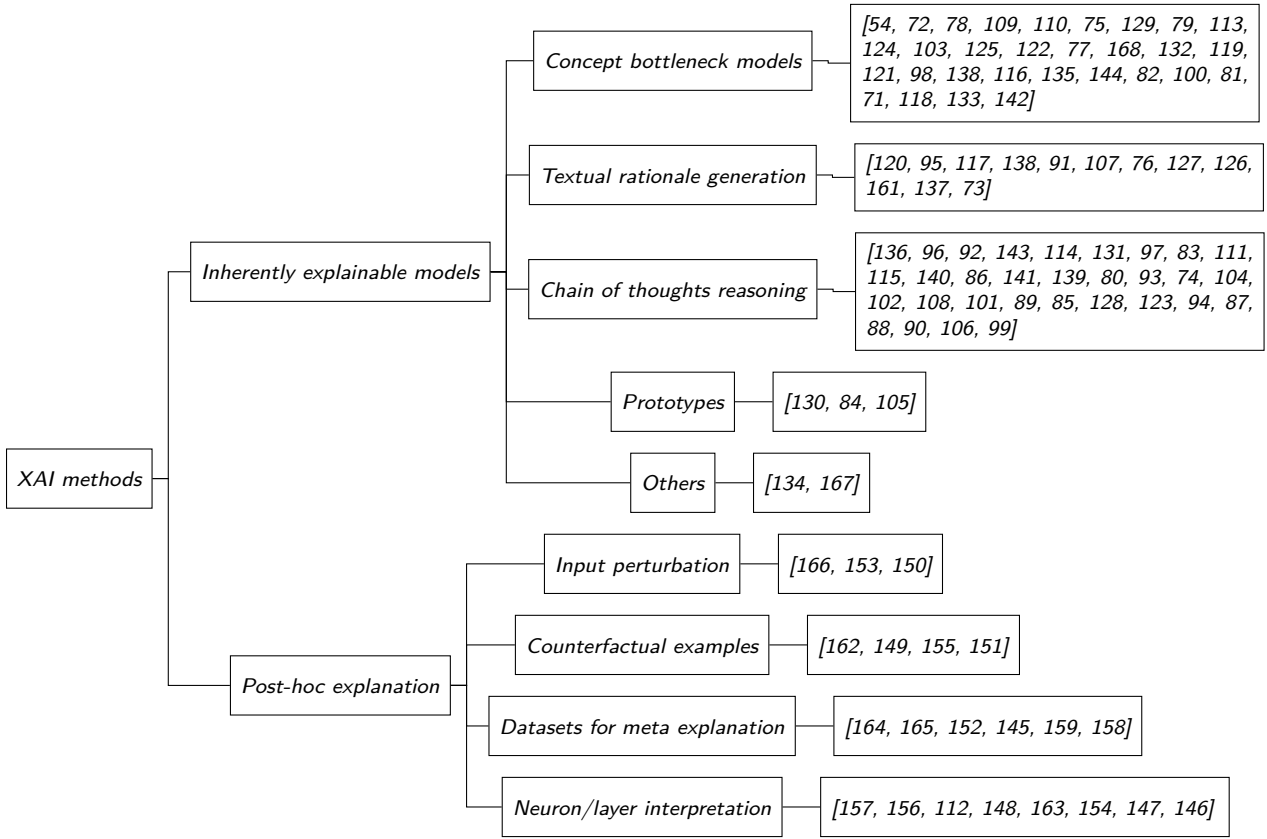
\begin{figure} 
    \begin{forest}                      
    forked edges,                        
    for tree={grow=0,draw, 
    font=\strut\scriptsize\sffamily\em},     
    [XAI methods
        [Post-hoc explanation
                    [Neuron/layer interpretation
                        [\parbox{0.3\linewidth}{\scriptsize \cite{wang2023reprompt,wen2024hard,li2023self,gandikota2023concept,shaham2024multimodal,kalibhat2023identifying,oikarinen2022clip,balasubramanian2024decomposing}}]]
                    [Datasets for meta explanation
                        [\scriptsize \cite{chegini2023identifying,gao2024lora,jain2022distilling,arias2023assessing,liu2024organizing,luo2025llm}]]
                    [Counterfactual examples
                        [\scriptsize \cite{luo2023zero,farid2023latent,kim2023grounding,augustin2022diffusion}]]
                    [Input perturbation
                        [\scriptsize \cite{knab2024dseg,sun2024explain,blucher2024decoupling}]]]
        [Inherently explainable models
                [Others
                    [\scriptsize \cite{kim2024vision,dombrowski2024trade}]]
                [Prototypes
                    [\scriptsize \cite{wan2024interpretable,mannix2024scalable,ando2023interpretable}]]
                [Chain of thoughts reasoning
                    [\parbox{0.3\linewidth}{\scriptsize \cite{liang2024toa,han2024image,han2024dme,wu2024voronav,kassab2023language,chen2024spatialvlm,zhang2023ecenet,ZhangLearning,chen2023reason,lian2023llm,rose2023visual,harvey2023visual,shao2024visual,meng2024few,liu2024chartthinker,grover2024navigating,zhao2024bba,kil2024ii,hwang2024safe,mondal2024kam,hu2024relax,fu2023dynamic,ge2023chain,qi2023art,nie2023reason2drive,ma2023dolphins,chen2023measuring,li2024cpseg,zheng2023ddcot,fu2023interpretable,uehara2024advancing}}]]
                [Textual rationale generation
                    [\parbox{0.3\linewidth}{\scriptsize \cite{zhu2023multimodal,xu2023drivegpt4,lai2024towards,wang2023vamos,wang2023decap,lin2024towards,achille2024interpretable,qi2024sniffer,banerjee2024learning,ku2023viescore,natarajan2024vale,libeyond}}]]
                [Concept bottleneck models
                    [\parbox{0.3\linewidth}{\scriptsize \cite{chen2023stair,fumanal2023artxai,liu2023chatgpt,oikarinen2023label,yang2023language,kim2023concept,panousis2023sparse,ren2024chatgpt,yan2023learning,yan2023robust,panousis2023hierarchical,patricio2023towards,xu2023retrieval,cui2023ceir,echterhoff2024driving,bhalla2024interpreting,wu2024concept,norrenbrock2024q,thakur2022multimodal,wang2023vamos,chiquier2024evolving,moayeri2023text,bie2024xcoop,kazmierczakclip,yow2024extract,menon2022visual,choi2024adaptive,hutowards,vandenhirtzstochastic,srivastavavlg}}]]]]
    \end{forest}
    \caption{Summary of the XAI methods presented in our study.}
    \label{taxonomy_general}
\end{figure}

\section{XAI Methods for Pretrained Foundation Models} \label{PFM_for_XAI}

\remi{In this section, we present and discuss different categories of PFMs for XAI. 
They are divided into two main groups. First, we examine inherently explainable models, which are designed to produce explanations by incorporating interpretable components directly into their architecture. Second, we explore post hoc methods, which encompass any external tool to the model, used after the training phase to provide explanations.}
\subsection{Inherently explainable models} \label{sec:inherently_interpretable_methods}

The complete list of inherently interpretable methods is presented in Table \ref{xai-methodsPFMforXAI}. Each method is associated with the taxonomy of \cite{speith2022review}, which includes scope, output format, functioning, and type of result.

\subsubsection{Concept bottleneck models} \label{sec:cbms}

\paragraph{Definition}

\remibis{A family of ante-hoc explainability techniques designed to make the predictions of DNNs interpretable by modifying their structure. Specifically, a DNN is partitioned into two components: (1) a concept predictor that maps input data to a set of semantic concepts, and (2) a classifier that predicts the final output class based on these concepts, as shown in Table \ref{fig:schema_CBM}. The concepts typically represent high-level, human-understandable features that are relevant to the prediction task. Explanations are derived from the latent space of concepts, as the model's predictions can be attributed to a specific set of concepts it focuses on during inference.}

\begin{figure}[ht]
    \centering
    \includegraphics[width=0.8\textwidth]{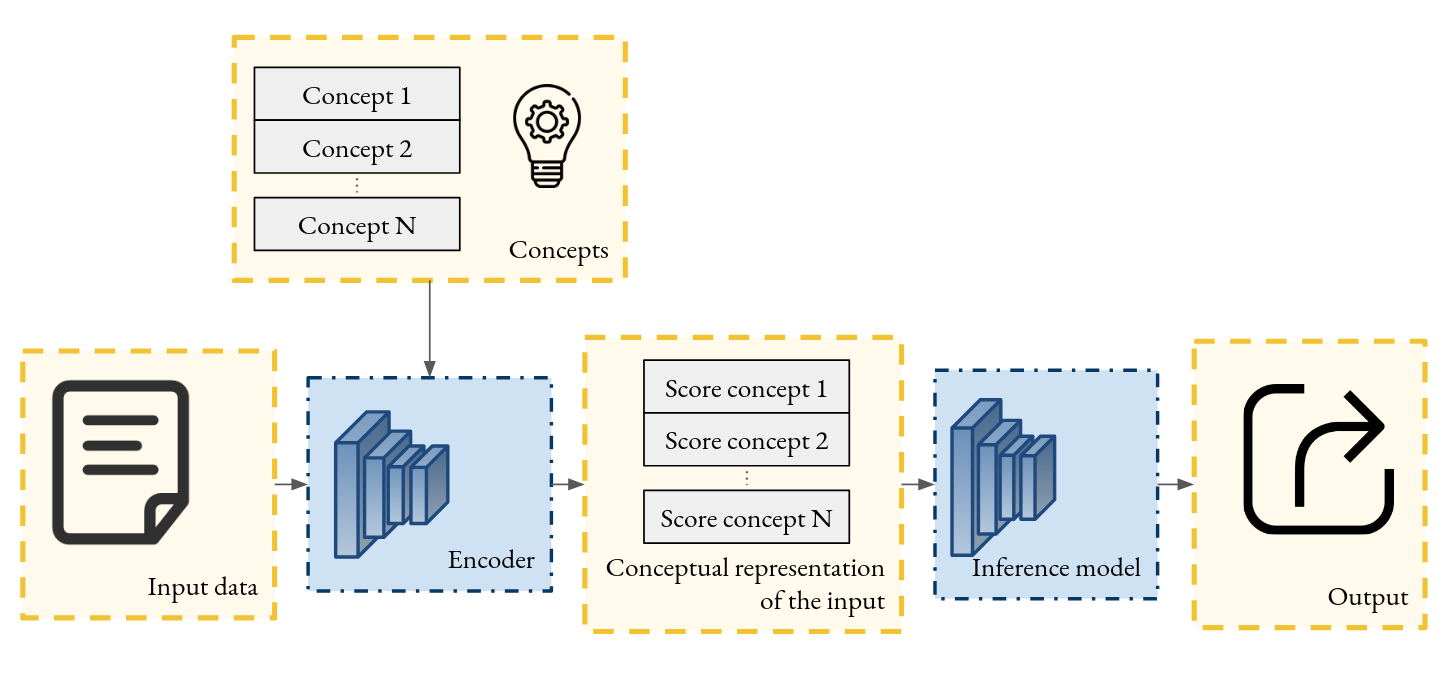}
    \caption{\remi{\textbf{Scheme of the principle of an inherently interpretable model through a concept bottleneck.} Given an input, the CBM first generates a conceptual representation based on a predefined set of concepts. Subsequently, the model produces an output using this conceptual representation.}}
    \label{fig:schema_CBM}
\end{figure}

\paragraph{Background}

\remi{Concept Bottleneck Models (CBMs) represent an interpretable approach in machine learning, where models make predictions based on high-level, human-interpretable concepts extracted from the data—often designated by descriptive terms (e.g.,  words), rather than on direct data-to-prediction mappings. While the term ``Concept Bottleneck Models'' is relatively new, this paradigm has roots in pre-deep learning literature \cite{kumar2009attribute, lampert2009learning}. The concept of CBMs was formally introduced by \cite{koh2020concept}, with similar ideas presented as ``Semantic Bottleneck Networks'' by \cite{losch2019interpretability}. Recent advancements have leveraged large language models to construct concepts from CLIP text embeddings \cite{yang2023language, oikarinen2023label}, giving rise to a family of CLIP-based CBMs. This direction has spurred extensive research \cite{chen2023stair, fumanal2023artxai, liu2023chatgpt, kim2023concept, panousis2023sparse}.}

\paragraph{Use of Pretrained Foundation Models}

\remi{In their inner functioning, CBMs require models to incorporate semantically meaningful concepts, which is challenging without specially tailored datasets. The introduction of PFMs has made it possible to overcome this limitation through their multimodal generalizability. By leveraging PFMs in the encoder, CBMs can effectively embed meaningful concepts. One of the pioneering works in this direction is \cite{menon2022visual}, which employs text descriptions of classes to augment zero-shot classification by CLIP through score thresholds on these descriptors. Additionally, \cite{liu2023chatgpt} uses CLIP to encode both image and text tokens associated with these concepts, resulting in a latent space that captures the combined encoding of text and image representations. An additional advantage of this approach is its training efficiency: since concept representations are inherently semantically meaningful and low-dimensional, training can focus primarily on the inference model, keeping it lightweight. \remibis{Beside CLIP, a noticable work is \cite{srivastavavlg} that uses Grounding DINO \cite{liu2025grounding} to spot the position of the detected concepts as bounding boxes.}}

\paragraph{Application and benchmark}

\remi{CBMs are primarily applied to image classification tasks, as the structure of the latent space—often aligned with words due to its design—naturally supports classification processes. Notably, CBMs have also been adapted for applications beyond image classification, including video understanding \cite{wang2023vamos} and image representation learning \cite{cui2023ceir}. Given the broad range of applications, a variety of datasets are utilized, spanning domains like medical imaging \cite{kim2023concept,yan2023robust}, art \cite{fumanal2023artxai}, and autonomous driving \cite{echterhoff2024driving}.}

\paragraph{Evaluation}

\remi{Primarily due to the challenges of conducting evaluations, many studies rely on qualitative explanations using datasets adapted to their respective models. However, some works have adopted tailored XAI evaluation metrics to provide quantitative insights. For example, deletion metrics \cite{petsiuk2018rise} are utilized in \cite{kazmierczakclip,panousis2023hierarchical}, while sparsity metrics \cite{chalasani2020concise} are employed in \cite{panousis2023sparse}.}

\subsubsection{Textual rationale generation}

\paragraph{Definition}

\remibis{A family of ante-hoc explainability techniques designed to make the predictions of DNNs interpretable by modifying their structure. These methods incorporate a specialized \remibis{componant} trained to generate textual justifications for the model's predictions. This \remibis{componant} produces explanations by decoding the latent space of the network (as illustrated in Table \ref{fig:schema_rationale}) \remibis{and often derived from an LLM}. The generated explanations typically take the form of concise textual statements that directly address the question: ``Why did the model make this specific inference?''.}

\begin{figure}[ht]
    \centering
    \includegraphics[width=0.7\textwidth]{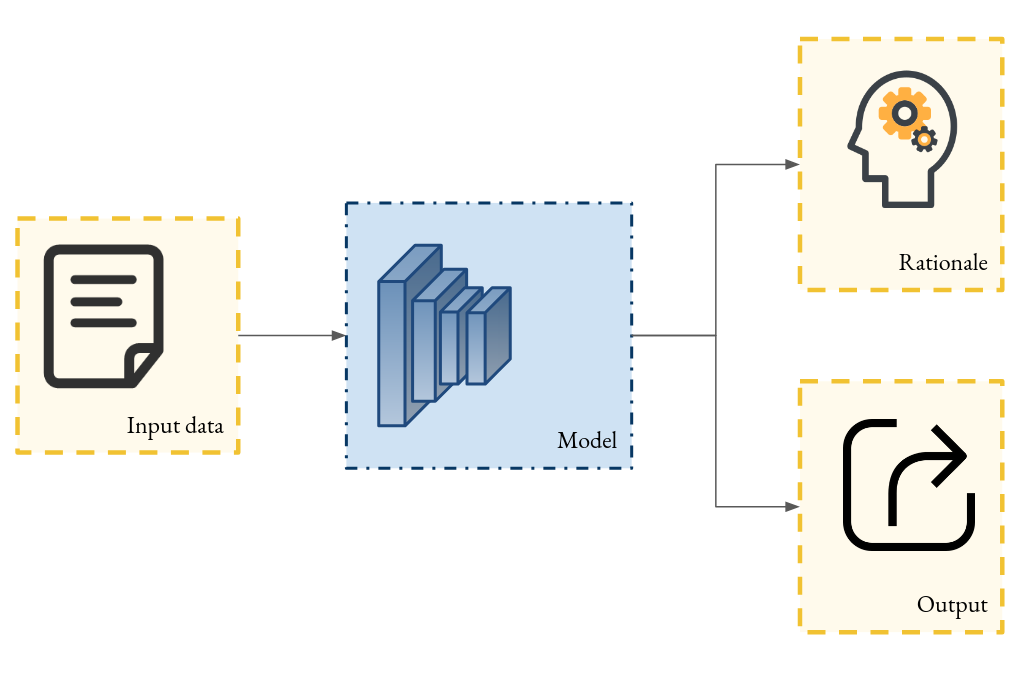}
    \caption{\remi{\textbf{Scheme of the principle of an inherently interpretable model through the generation of textual rationale.} Given an input, the model generates a textual rationale alongside its output, providing insights into the reasoning behind the prediction. Compared to other methods that imply some interpretable decomposition of the model, the model here can be opaque.}}
    \label{fig:schema_rationale}
\end{figure}

\paragraph{Background}

\remi{Rationales provide context and reasoning that closely align with human language and cognitive processes, offering explanations that are more naturally interpretable by humans. This concept has been explored since the early days of AI, often through template-based approaches \remibis{that consists in triggering predefined sentences based on specific conditions within the system framework being met \cite{shortliffe1975model,van2004explainable}}. However, generating multimodal, text-based rationales from deep vision models has been particularly challenging due to the complexity of combining visual and linguistic information. The first successful approach to tackle this was introduced by \cite{park2018multimodal}, which involved training a model on a custom dataset specifically designed for multimodal rationale generation. Recent advancements in large language models \cite{brown2020language} and large multimodal models \cite{li2023blip} have since opened new avenues for rationale generation, reducing dependence on dataset-specific constraints and enabling more flexible and generalizable methods across the field.}

\paragraph{Use of Pretrained Foundation Models}

\remi{Analogous to Concept Bottleneck Models, the emergence of multimodal PFMs has significantly accelerated the development of approaches utilizing textual rationales. Due to their more flexible framework, a wider variety of PFMs have been employed in this domain, including CLIP \cite{wang2023decap, lai2024towards}, as well as BLIP \cite{zhu2023multimodal} and LLaVA \cite{achille2024interpretable}. This diversity is particularly noteworthy since the latter two models are specifically designed for image captioning tasks, highlighting their adaptability and relevance in generating coherent and contextually appropriate textual explanations. In addition, we also noticed some works using extra modules to guide the image captioning module on areas the model focuses on \cite{natarajan2024vale,nguyen2024langxai}.}

\paragraph{Application and benchmark}

\remi{The established nature of rationale-generating methods with standardized architectures has paved the way for adaptations tailored to specific domains, such as autonomous driving \cite{xu2023drivegpt4} and harmful meme detection \cite{lin2024towards}. Recent work has also focused on refining architectures to enhance informativeness \cite{wang2023decap} and contextual relevance \cite{zhang2023ecenet}, and on extending rationale generation to more complex data types, such as video \cite{wang2023vamos}. Additional approaches, such as \cite{lai2024towards}, incorporate visual explanations through bounding boxes, while others like \cite{ku2023viescore} apply rationales to image quality assessment. The VQA-X dataset \cite{park2018multimodal} is a popular benchmarking tool, providing a standard for comparison across studies in rationale-based explainability.}

\paragraph{Evaluation}

\remi{The field of rationale generation benefits from well-established text perceptual similarity metrics \cite{papineni2002bleu,lin2004rouge}, providing a solid baseline for evaluating generated explanations. The standard approach for assessing explanation quality involves calculating the similarity between generated rationales and ground-truth rationales found in datasets like VQA-X \cite{park2018multimodal}. Furthermore, more specialized datasets are available for domain-specific applications, such as \cite{kim2018textual}, which is tailored for autonomous driving contexts.}

\subsubsection{Chain of thought reasoning} \label{CoT_PFM_for_XAI}

\paragraph{Definition}

\remibis{A family of ante-hoc explainability techniques designed to make the predictions of DNNs interpretable by modifying their \remibis{way to produce inference}. This involves decomposing the network into multiple interpretable blocks, as illustrated in Table \ref{fig:schema_cot}. Chain-of-thought explanation aims to elucidate the reasoning process that leads to the model's predictions. The interpretability stems from the fact that humans can more easily understand and follow the sequential reasoning behind the decision. Such methodologies are frequently implemented in large language models to enhance the transparency of their decision-making processes.}

\begin{figure}[ht]
    \centering
    \includegraphics[width=0.8\textwidth]{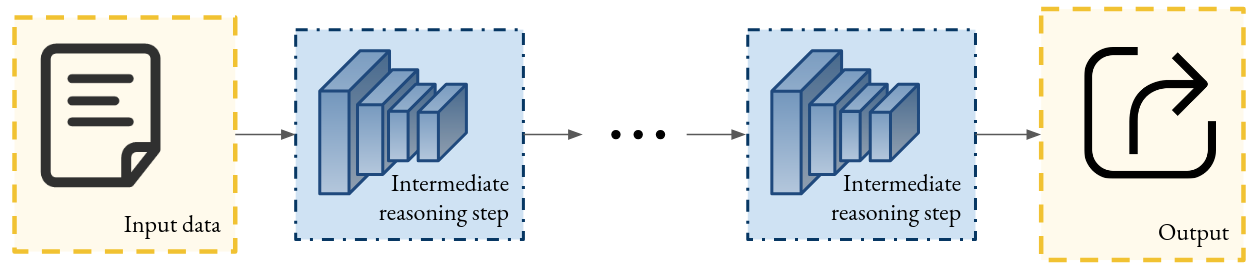}
    \caption{\remi{\textbf{Scheme of the principle of an inherently interpretable model through chain of thought.} The model processes its input in multiple sequential steps, resulting in an iterative reasoning process.}}
    \label{fig:schema_cot}
\end{figure}

\paragraph{Background}

\remi{The initial developments of Chain of Thought (CoT) explanations originated in language-only tasks with large language models (LLMs) \cite{wei2022chain}. These methods, often referred to as zero-shot approaches, involve sequentially applying prompts to simulate a reasoning process closer to human-like thinking. In this paradigm, the explanation unfolds as a series of decomposed steps or actions (see Figure \ref{fig:schema_cot}). A related paradigm, known as few-shot CoT, extends this by using a parallel decomposition of reasoning steps. With advancements in PFMs and their capability to process multimodal inputs, CoT reasoning has now been adapted for visual tasks \cite{zheng2023ddcot,chen2023measuring}. Additionally, increasingly complex reasoning structures have been developed to enhance outputs through advanced integration of reasoning blocks \cite{liu2024chartthinker,qi2023art,hwang2024safe}.}

\paragraph{Use of Pretrained Foundation Models}

\remi{Due to their close relationship with large language models, CoT techniques have naturally extended to multimodal PFMs. Notably, BLIP-2 has become particularly popular in CoT applications \cite{nie2023reason2drive,chen2023reason,wu2024voronav}. Additionally, models like GPT-4 are increasingly used to develop CoT-based techniques \cite{ZhangLearning}, especially as API-based prompting approaches simplify integration. }

\paragraph{Application and benchmark}

\remi{Building on a solid foundation from previous work in large language models, CoT-based methods in XAI have gained substantial traction in the literature. These approaches leverage step-by-step reasoning to improve both explainability and accuracy. Researchers have explored a variety of CoT-inspired enhancements, such as Socratic reasoning \cite{qi2023art}, visual/non-visual information separation \cite{hu2024relax}, the integration of knowledge graphs \cite{mondal2024kam}, decision tree frameworks \cite{fu2023interpretable}, and segmentation techniques \cite{li2024cpseg}. Contrary to the common assumption that interpretability often reduces model accuracy, CoT methods have shown promise in maintaining or even improving accuracy when integrated with PFMs. This versatility has also led to the development of CoT approaches for specialized domains, including autonomous driving \cite{hwang2024safe,nie2023reason2drive,ma2023dolphins,zheng2023ddcot} and mathematical reasoning \cite{qi2023art,zhao2024bba}.}

\paragraph{Evaluation}

\remi{Given the complexity of capturing a ground truth that accurately reflects human reasoning, evaluating CoT methods is substantially more challenging than assessing simpler textual rationales. As a result, many studies rely on qualitative examples for evaluation, as seen in works such as \cite{zheng2023ddcot,hu2024relax,ge2023chain}. In specific cases, researchers have devised evaluation methods involving sub-questions that aim to break down reasoning processes, sometimes even proposing these methods as potential benchmarks, as seen in \cite{chen2023measuring,nie2023reason2drive}.}

\subsubsection{\remi{Prototypical networks}} \label{sec:prototypes}

\paragraph{Definition}

\remibis{A family of ante-hoc explainability techniques designed to make the predictions of DNNs interpretable by modifying their structure. Specifically, the DNN is divided into three core components (Table \ref{fig:schema_prototypes}): (1) an encoder that transforms inputs into fixed-size vectorized representations, (2) a mapper that translates the latent space vectors into semantic prototypes, and (3) an inference module that derives the final output from these prototypes. Unlike CBMs, this approach learns the set of prototypes during training. Interpretability arises from attributing the model's predictions to specific prototypes activated during inference, offering insight into the decision-making process. These techniques can be applied to any model. }

\begin{figure}[ht]
    \centering
    \includegraphics[width=0.8\textwidth]{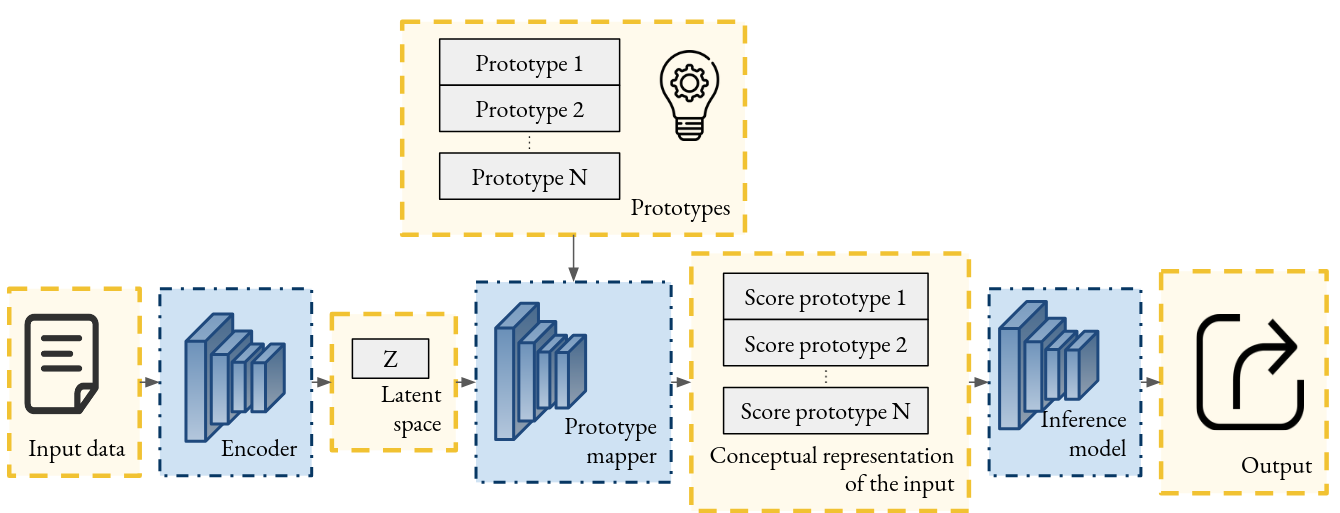}
    \caption{\remi{\textbf{Scheme of the principle of an inherently interpretable model through prototypes.} The input is embedded into a latent space and mapped to regions corresponding to previously learned prototypes, which are then used to produce the output.}}
    \label{fig:schema_prototypes}
\end{figure}

\paragraph{Background}

\remi{Prototypical networks were first introduced by \cite{snell2017prototypical} for few-shot and zero-shot learning, where the concept of prototypes is closely related to clustering techniques. This idea was later adapted for interpretable object recognition, such as in the work on ProtoNet \cite{chen2019looks}, which probes \remibis{training images of each class} to identify common prototypes. Subsequent improvements have incorporated methods like decision trees \cite{nauta2021neural} and vision transformers \cite{xue2022protopformer} to enhance model performance. However, these approaches still face challenges related to computational costs. To address these limitations, new solutions leveraging Prototypical Networks with PFMs have been proposed, offering more scalable and interpretable alternatives \cite{wan2024interpretable,mannix2024scalable,ando2023interpretable}.
}

\paragraph{Use of Pretrained Foundation Models}

\remi{One of the key benefits of using PFMs in prototypical networks is the ability to represent prototypes across multiple modalities, such as text using CLIP, while simultaneously reducing both computational and labeling costs \cite{wan2024interpretable}. Additionally, the incorporation of models like DINOv2 \cite{oquab2023dinov2} and SAM \cite{zou2024segment}, which enable the segmentation of relevant regions within an image, has been shown to enhance the identification of more expressive prototypes \cite{mannix2024scalable,ando2023interpretable}.}

\paragraph{Application and benchmark}

\remi{The applications of prototypes using PFMs are diverse. \cite{wan2024interpretable} and \cite{mannix2024scalable} applied their methods to the CUB dataset \cite{wah2011caltech}, which has been widely used in previous research on prototypical networks. Additionally, \cite{ando2023interpretable} propose the application of prototypical networks to medical images.}

\paragraph{Evaluation}

\remi{Current methods are often limited to qualitative examples, as the prototype set is not fixed by the dataset, which complicates the evaluation of explanations. However, \cite{wan2024interpretable} addresses this challenge by proposing a quantitative evaluation using the deletion metric \cite{petsiuk2018rise}, which assesses the importance of the pixels highlighted by their explanations.}

\subsubsection{Others}

\remi{Among the corpus of inherently explainable methods, there are certain approaches that are too specific to fit neatly into the families previously discussed. For example, Finetune \cite{dombrowski2024trade} presents a method for fine-tuning diffusion models to enhance their interpretability. This is achieved by using CLIP to generate activations related to textual concepts, specifically applied to radiology images. Similarly, sViT \cite{kim2024vision} employs SAM (Segment Anything Model) to segment the input image, allowing for the clustering of the image into semantically meaningful regions, as opposed to the typical patch-based process used in vision transformers.}

\subsection{Post-hoc \remi{explanation methods}} \label{sec:post_hoc_methods}

The complete list of post-hoc methods is presented in Table \ref{xai-methods-posthoc}. Each method is associated with the taxonomy of \cite{speith2022review}, which includes scope, output format, functioning, and type of result.

\subsubsection{Input perturbation}

\paragraph{Definition}

\remibis{A family of post-hoc explainability techniques designed to make an explanation of any model without modifying the structure of the model by probing its behavior on a set of perturbed input variants. The process is typically divided into two steps (Table \ref{fig:schema_perturbating_input}): (1) generating outputs for various perturbations of the input, and (2) training an auxiliary interpretable model to approximate the local behavior of the original model around the perturbed input space. Explanations are then derived by analyzing this local approximation model, focusing on the impact of each input feature on the model's decision (feature attribution). These techniques can be applied to any model.}

\begin{figure}[ht]
    \centering
    \includegraphics[width=0.8\textwidth]{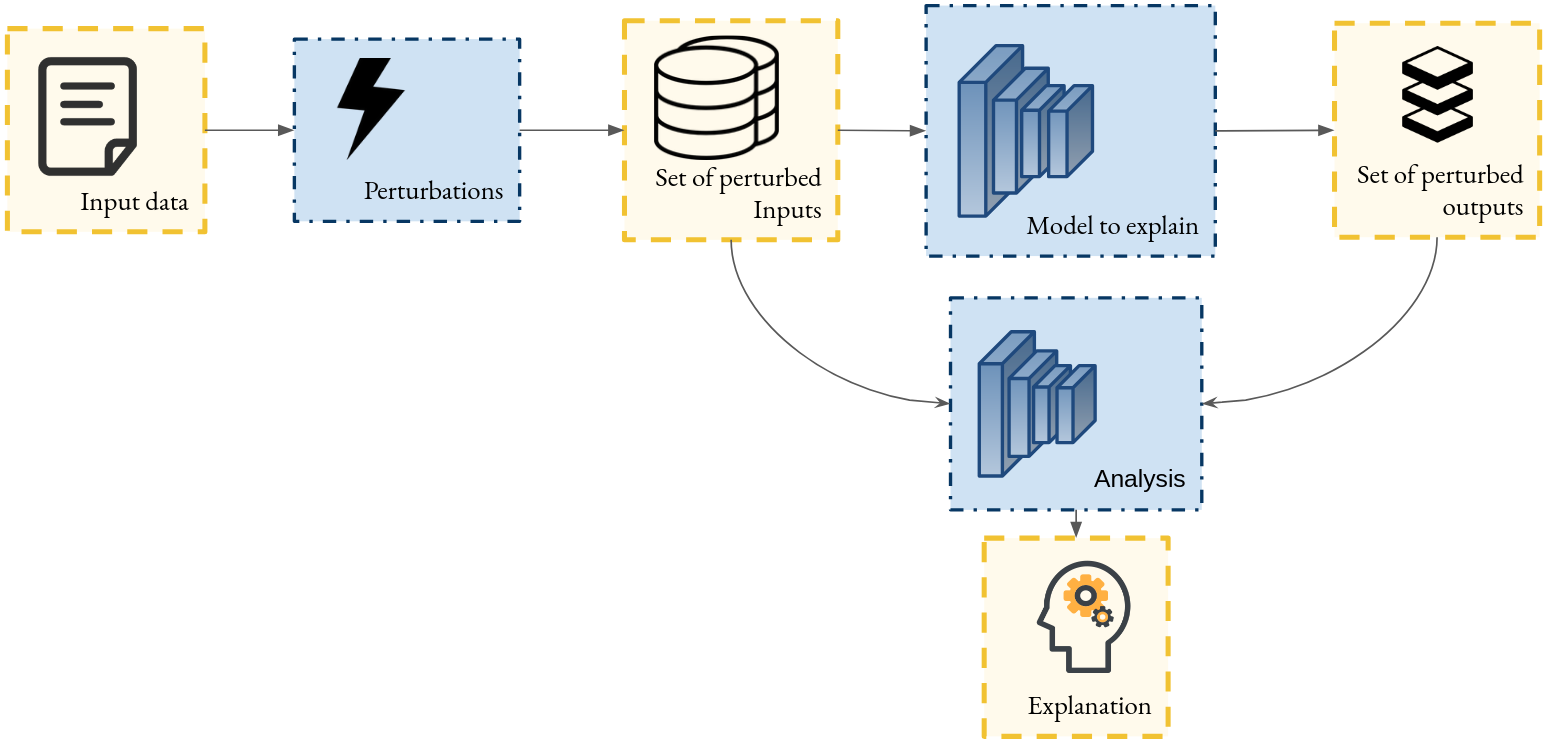}
    \caption{\remi{\textbf{Scheme of the principle of post-hoc explanation by perturbing input data.} Given an input, a set of perturbed samples is generated. The model's behavior in response to these perturbations is then analyzed, and the resulting analysis provides explanations for the model's inference.}}
    \label{fig:schema_perturbating_input}
\end{figure}

\paragraph{Background}

\remi{Input perturbation refers to a class of post-hoc techniques that aim to explain the inference of a given model by testing the model on a set of slightly perturbed variants of the input sample. This approach has gained significant interest, as the diversity of DNN architectures necessitates a flexible and robust process for producing sample-wise explanations. Early contributions to this field include methods like LIME \cite{lime} and SHAP \cite{lundberg2017unified}, which have become widely adopted for their ability to provide interpretable explanations for opaque models. Subsequent works have focused on improving these methods, particularly by developing better approximations for high-dimensional inputs, such as images.}

\paragraph{Use of Pretrained Foundation Models}

\remi{Given the challenges associated with creating superpixels, PFMs dedicated to semantic segmentation, such as SAM \cite{zou2024segment}, have proven useful for producing semantically meaningful decompositions of images. Consequently, extensions of SHAP have been developed using SAM \cite{blucher2024decoupling, sun2024explain}, which enhance the quality of explanations. In addition, \cite{knab2024dseg} have proposed integrating semantic segmentation with LIME.}

\paragraph{Application and benchmark}

\remi{Due to the versatility of these methods, there are no specific datasets or applications that are strictly tied to them. While the papers presented here often focus on commonly used datasets like ImageNet or COCO, these techniques can be adapted to a wide range of datasets and tasks.}

\paragraph{Evaluation}

\remi{In the context of attribution-based XAI methods, a variety of quantitative evaluation metrics are commonly employed. \cite{sun2024explain} and \cite{knab2024dseg} adopt the deletion metric \cite{petsiuk2018rise} to assess the trustworthiness of explanations, which measures the drop in model accuracy when the most important pixels are occluded. This approach helps evaluate how crucial the identified features are to the model’s decision-making process. Additionally, \cite{blucher2024decoupling} introduce a variant of this metric, named as Symmetric Relevance Gain. Note that other metrics, such as Noise Stability \cite{zhang2021xai} and Preservation Check \cite{goyal2019counterfactual}, are also used \cite{knab2024dseg}.}

We focus here on post-hoc methods from our corpus of articles. The full list of methods is presented in Table~\ref{xai-methods-posthoc}. Notably, we associate each method with the taxonomy of \cite{speith2022review}, including scope, output format, functioning, and type of result. 

\subsubsection{Counterfactual examples}

\paragraph{Definition}

\remibis{A family of post-hoc explainability techniques designed to make an explanation of any model without modifying the structure, by searching for counterfactual examples. To do so, an optimizing process is performed to search for a minimal perturbation that induces high changes in the prediction (for example a label shift in classification tasks), represented as $do(Z=z_0+\varepsilon)$, where $z_0$ is a fixed input value and $\varepsilon$ is a small perturbation, using a common notation in causal inference \cite{peters2017elements}. A figure representing the process is available in Table \ref{fig:schema_counterfactual}. The resulting explanation is the pair constituted of a original image, the perturbated image, and the perdiction given the original image and the prediction given the perturbated image. Explanations are derived from the fact that finding counterfactuals give to the users examples of causal interventions that rules the functioning of the model. These methods can be applied to any model.}

\begin{figure}[ht]
    \centering
    \includegraphics[width=0.8\textwidth]{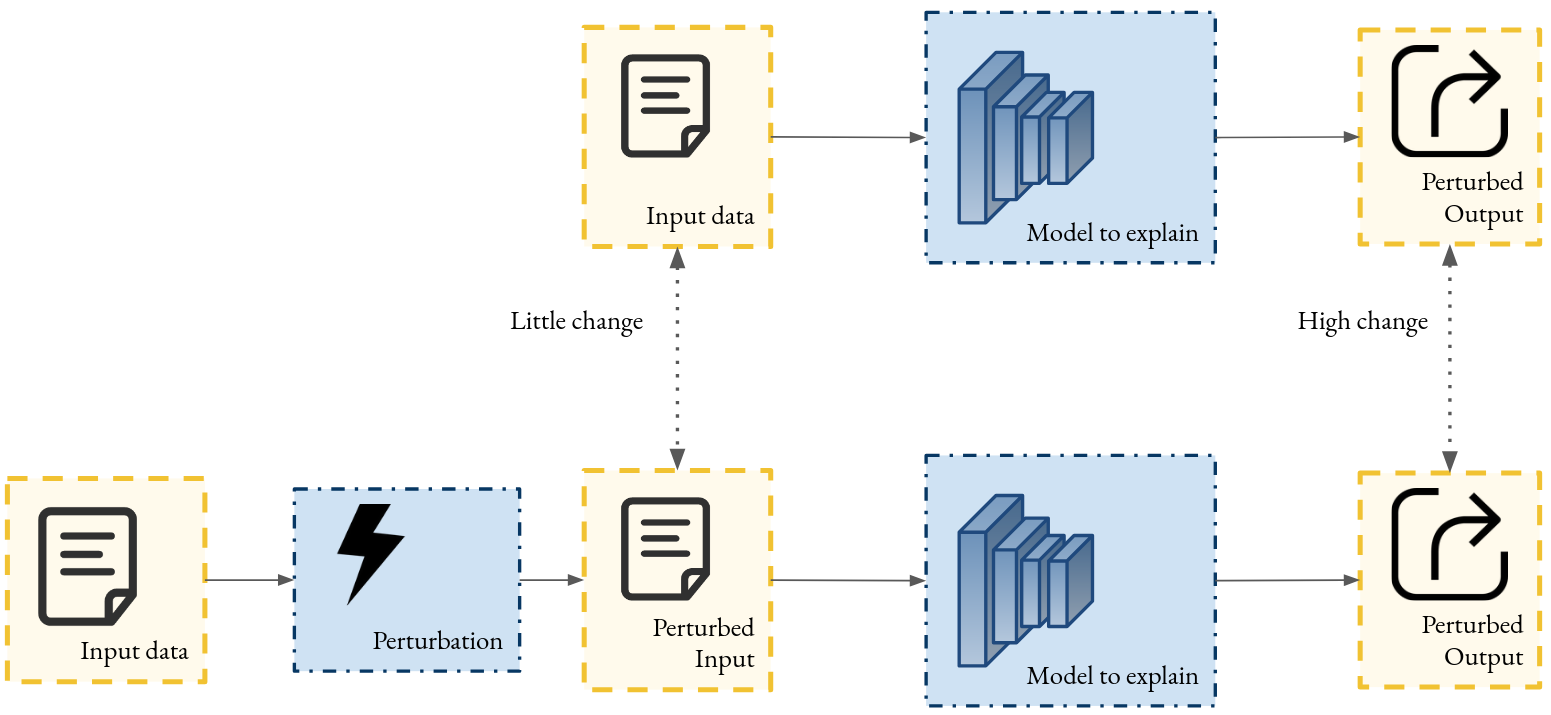}
    \caption{\remi{\textbf{Schema of the principle of post-hoc explanation through counterfactuals.} A minimal perturbation is applied to the input to generate a variant that results in a significant change in the model's inference compared to the original input.}}
    \label{fig:schema_counterfactual}
\end{figure}

\paragraph{Background}

\remi{Counterfactual explanations seek to identify minimal perturbations in the input data that induce a significant shift in the model’s prediction. These methods have relevance across disciplines, including philosophy, psychology, and social sciences, where theories on counterfactual reasoning have been extensively explored \cite{pearl2009causal,woodward2005making}. Recent advancements in machine learning, particularly those that enable the encoding of data into structured latent spaces, have catalyzed a new wave of counterfactual generation techniques. A foundational contribution in this domain is provided by \cite{pawelczyk2020learning}, which leverages variational encoders to create counterfactual instances, initially focusing on tabular data. This approach has been further refined and generalized; for instance, \cite{duong2023ceflow} integrates flow-based models to enhance the flexibility of counterfactual generation. In computer vision, latent space manipulation for counterfactual creation has also gained traction, as seen in the methods proposed by \cite{zhao2023counterfactual} and \cite{yang2021generative}.}

\paragraph{Use of Pretrained Foundation Models}

\remi{For image inputs, the generation of counterfactuals heavily relies on advancements in image editing techniques. In this regard, diffusion models like Stable Diffusion \cite{rombach2021high} have shown significant promise. By optimizing for gradients within the diffusion steps that highlight model sensitivities, these methods can effectively generate meaningful counterfactuals \cite{farid2023latent,augustin2022diffusion}. Another approach leverages PFMs to identify counterfactual directions that align with specific concepts. For instance, \cite{kim2023grounding} utilize the CLIP embedding space to discover directions corresponding to concept addition in a model-agnostic manner \remibis{(note that CLIP-QDA also proposes  counterfactuals but is not model-agnostic)}. Similarly, \cite{luo2023zero} apply CLIP to interpret the latent space of StyleGAN, producing counterfactuals in the form of edited images.}

\paragraph{Application and benchmark}

\remi{To date, most methods in this domain have been demonstrated on well-known datasets like CELEB-A \cite{lee2020maskgan} and CUB \cite{wah2011caltech}, providing convenient, straightforward use cases. These datasets support methods by offering controlled scenarios with interpretable attributes, facilitating analysis and comparison. In terms of application tasks, the primary focus has been on image classification. However, these approaches hold potential for adaptation to a range of tasks involving image inputs, suggesting the feasibility of extending these techniques to other domains in visual processing \remibis{like regression or semantic segmentation.}}

\paragraph{Evaluation}

\remi{A key criterion for effective counterfactual generation is maintaining proximity to the original input image, which ensures the counterfactual’s relevance and interpretability. Consequently, popular evaluation metrics for counterfactual quality are derived from image quality assessment frameworks, such as the Fréchet Inception Distance (FID) \cite{heusel2017gans} and the Learned Perceptual Image Patch Similarity \cite{zhang2018unreasonable}, where a smaller distance generally indicates a more effective counterfactual. Given the importance of perceptual similarity, traditional distance metrics like Peak Signal-to-Noise Ratio and Structural Similarity Index Measure are less commonly used in this context, as they may not capture \remibis{semantic} nuances as effectively.}

\subsubsection{Datasets for meta explanation} \label{dataset_bias_post_hoc}

\paragraph{Definition}

\remibis{A family of post-hoc explainability techniques designed to explain the global functioning of a model without modifying the structure by probing its responses on an auxiliary dataset specifically designed to reveal potential biases. Explanations are generated through statistical analysis of the model's behavior across the entire dataset (Figure \ref{fig:schema_meta_explanation}). These techniques can be applied to any model.}

\begin{figure}[ht]
    \centering
    \includegraphics[width=0.8\textwidth]{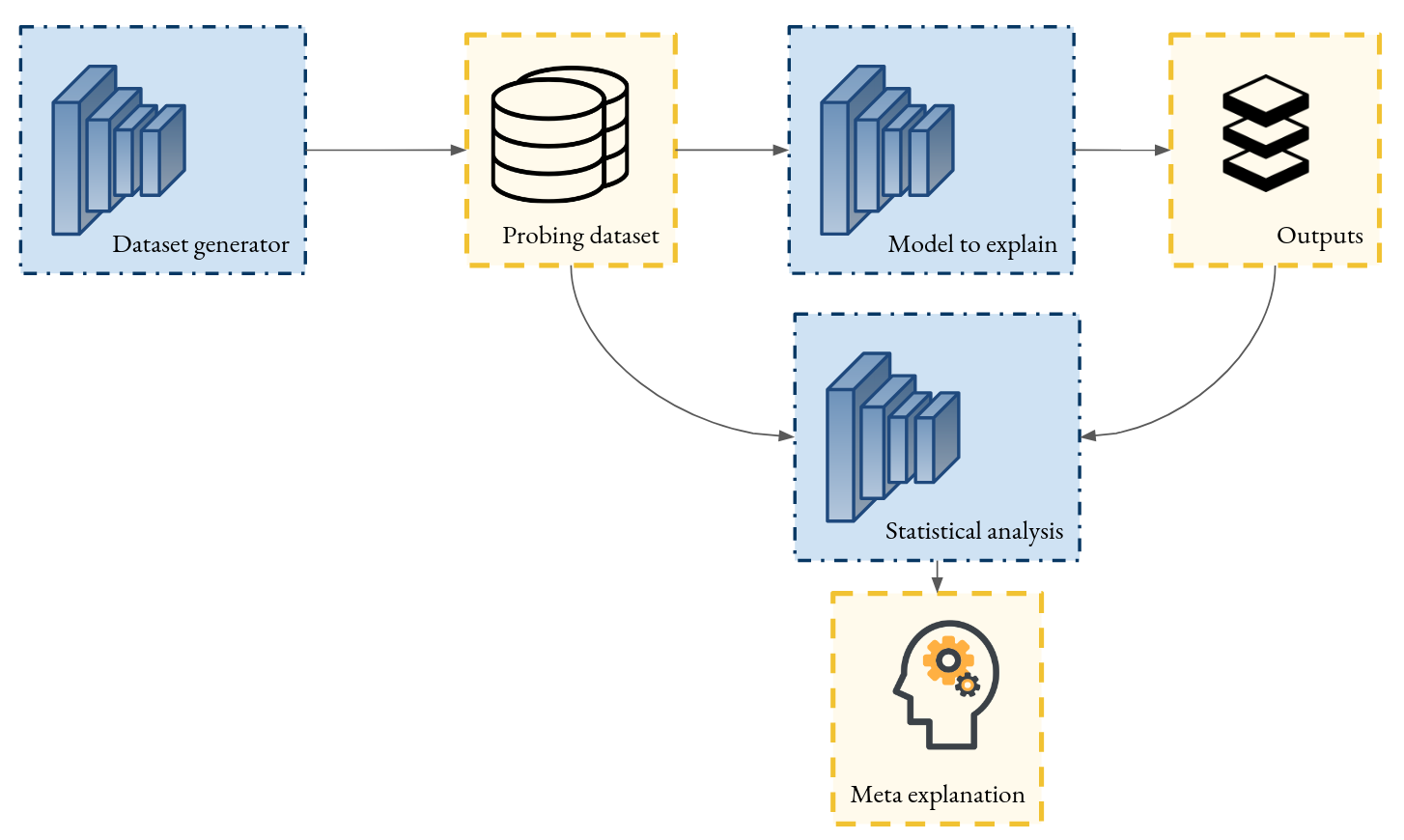}
    \caption{\remi{\textbf{Scheme of the principle of post-hoc explanation through meta explanations.} A dataset, generated through a prior process, is provided to the model. The model’s responses are then analyzed through statistical methods to produce a meta-explanation that offers insights into the model's behavior.}}
    \label{fig:schema_meta_explanation}
\end{figure}

\paragraph{Background}

\remi{The prevalence of biases in deep neural networks has driven substantial interest in studying model sensitivity to various forms of bias, with foundational work by \cite{szegedy2013intriguing} exploring this topic early on. This line of research spurred the creation of datasets specifically designed to probe for bias, either by extending existing datasets \cite{xiao2020noise} or through the development of entirely new ones \cite{franchi2022muad}. Such datasets enable statistical analyses of a model's behavior in response to biases, an approach sometimes referred to as ``meta-explanation'', e.g., in the review by Speith \cite{speith2022review}. However, due to the high costs associated with designing these datasets, progress has been limited. The advent of PFMs has mitigated some of these costs by facilitating artificial dataset creation for meta-explanation purposes, as illustrated by recent works in this area \cite{chegini2023identifying,gao2024lora,jain2022distilling,arias2023assessing}.}

\paragraph{Use of Pretrained Foundation Models}

\remi{As previously noted, a fundamental aspect of meta explanations is the design of a probing dataset. Here, PFMs offer significant value by generating high-quality images with flexible customization options. For instance, \cite{arias2023assessing} leverage stable diffusion to build a dataset featuring objects in varying contexts (e.g., with and without background), providing controlled conditions to study model biases. Similarly, \cite{gao2024lora} generate images that follow specific logical reasoning criteria, enhancing the robustness of bias analysis. Another advantage of PFMs is their ability to map both text and image data into a shared latent space, as demonstrated by \cite{jain2022distilling}, who use CLIP to represent model failures in this space as textual attributes, offering a more interpretable view of latent biases.}

\paragraph{Application and benchmark}

\remi{Currently, research on model biases in deep learning primarily targets the biases known to be particularly challenging, such as over-reliance on background features \cite{arias2023assessing}, societal biases that stem from dataset imbalances \cite{jain2022distilling}, and limitations in reasoning abilities \cite{gao2024lora}. Most of this work focuses on images representing everyday objects, including categories like food, transportation, and faces.}

\paragraph{Evaluation}

\remi{This type of explanation serves as a benchmark in itself, making direct comparisons between different meta-explanation techniques less meaningful. Instead, the evaluation of the quality of these methods is typically left to the discretion of the user, relying on qualitative examples to assess their effectiveness and relevance.}

\subsubsection{Neuron/layer interpretation} \label{sec:neur_interpret}

\paragraph{Definition}

\remibis{A family of post-hoc explainability techniques designed to explain the global functioning of a model without modifying the structure. These methods operate by identifying patterns that most strongly activate a specific neuron or layer within a DNN, as illustrated in Figure \ref{fig:schema_layer_interpret}. Two primary strategies are commonly employed. The first involves optimization techniques, where the objective is to determine an input \cite{wang2023reprompt, wen2024hard} or latent space direction \cite{li2023self, gandikota2023concept} that maximizes the activation of the targeted pattern. In this case, the explanation is represented by the optimized input obtained through this process. The second strategy leverages a dataset of images, selecting the top-activating examples \cite{shaham2024multimodal, kalibhat2023identifying}. Here, the explanation is provided by the subset of examples that elicit the strongest activations. These methods can be applied to any model.}

\begin{figure}[ht]
    \centering
    \includegraphics[width=0.6\textwidth]{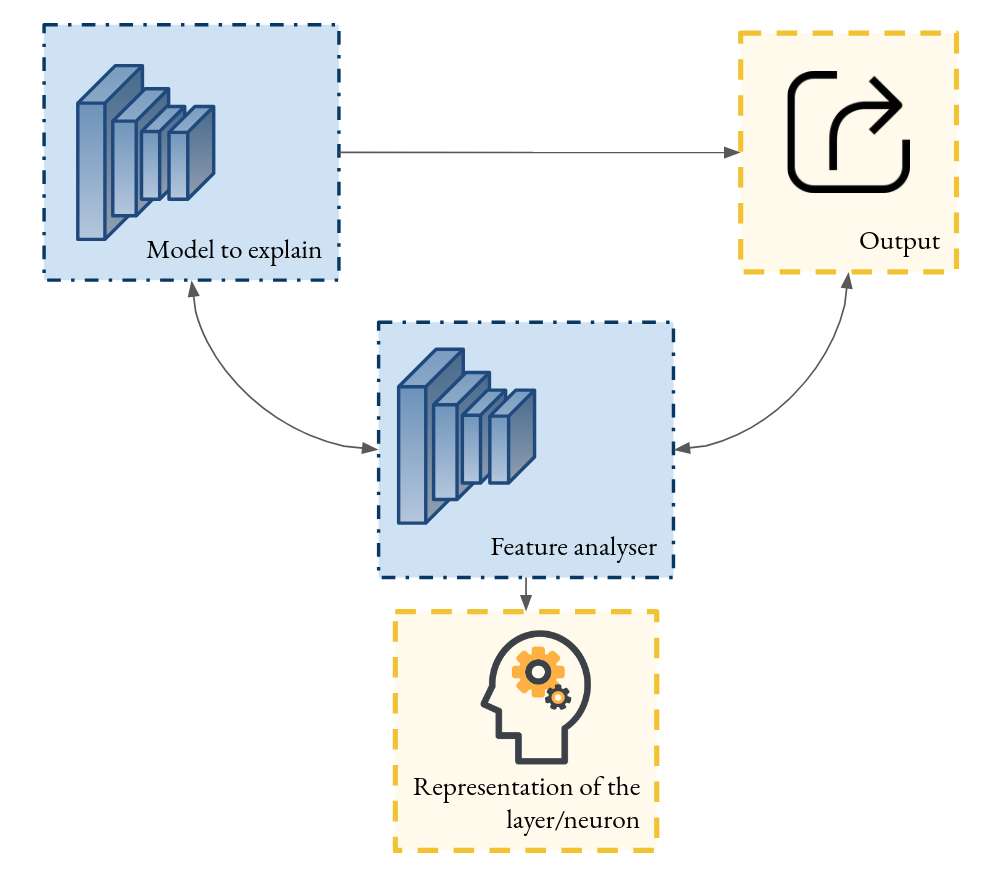}
    \caption{\remi{\textbf{Scheme of the principle of post-hoc explanation through neuron or layer interpretation.} Given a specific layer or neuron to analyze, an optimization module is employed to probe the layer's behavior. The output of this process provides insights into what the layer or neuron is sensitive to.}}
    \label{fig:schema_layer_interpret}
\end{figure}

\paragraph{Background}

\remi{As DNNs operate by extracting meaningful \remibis{patterns} to perform inference, a natural approach to understanding their behavior is to investigate which \remibis{patterns trigger a selected layer or neuron}. This type of analysis dates back to the earliest papers presenting foundational architectures, such as \cite{krizhevsky2012imagenet,zeiler2014visualizing}. Many approaches, often referred to as ``feature visualization'' \cite{olah2017feature}, aim to reveal to the user the specific \remibis{patterns} a network focuses on during its processing. Despite the increasing width and complexity of modern networks, which make interpretation more challenging, numerous works continue to explore methods for interpreting layers. Notably, areas such as text-based interpretations and the relationship between prompts and latent space have become central themes in recent research.}

\paragraph{Use of Pretrained Foundation Models}

\remi{Compared to traditional feature visualization methods, the integration of PFMs allows for a more advanced analysis, extending beyond merely displaying the inputs to which the model is sensitive. PFMs enable the connection of these inputs to other modalities, offering a richer interpretability framework. For instance, CLIP can be used to relate the top-activating images to textual concepts, as seen in works by \cite{oikarinen2022clip} and \cite{kalibhat2023identifying}. Additionally, CLIP enhances the optimization process in methods designed to interpret the relationship between prompts and generated images, particularly in text-to-image models \cite{wang2023reprompt, wen2024hard}.}

\paragraph{Application and benchmark}

\remi{In terms of task types, we have identified two main areas of application. The first one is text-to-image translation, where neuron/layer interpretation is of particular interest due to the significance of prompt engineering. A deeper understanding of the relationship between the input prompt and the network's behavior has substantial implications for the development of more effective \remibis{prompt} engineering techniques. The second type of task focuses on feature extraction, where the goal is to inspect a pretrained backbone model independently of the final output layer. Here, the focus is placed on understanding the behavior of specific portions of the network.}

\paragraph{Evaluation}

\remi{The evaluation of explanation quality is largely dependent on whether the method is applied to text-to-image generation or not. For image generation tasks, the quality of explanations often depends on the user's intent, making perceptual distances and user studies key evaluation metrics. These evaluations focus on how well the generated explanations align with the intended manipulation or interpretation of the image. In contrast, for methods that inspect the backbone of a pretrained model, quantitative evaluations are more common. These evaluations often involve detecting expected samples or specific \remibis{patterns} on toy examples, providing a more objective measure of the method's effectiveness in identifying meaningful network activations and features \cite{kalibhat2023identifying}.}

\section{Evaluating explanations in the era of PFMs} \label{XAI_evaluation}

\subsection{Required axioms for XAI} \label{Axioms_XAI}

Evaluating different XAI methods is important to measure their relative effectiveness. As pointed out by \cite{nauta2023anecdotal}, initial evaluations of XAI techniques predominantly relied on qualitative demonstrations, presenting obstacles in conducting unbiased comparisons across different methods.

To address this challenge, efforts have been made to identify key properties that define a good explanation, particularly in terms of audience comprehension. In the literature, the quality of an explanation depends on many factors, encouraging researchers to articulate various axioms (or desiderata) that characterize the attributes of a satisfactory explanation. Prior works have defined a set of axioms deemed most pertinent \cite{liao2022connecting,arrieta2020explainable}. Consequently, due to the inherent subjectivity involved, many qualifiers exist, resulting in diverse axiom lists corresponding to the multitude of papers addressing this question.
Explanation quality is inherently subjective, with different users prioritizing various aspects based on their preferences and needs. For instance, while one user may value \textit{correctness}, ensuring the explanation accurately reflects the model's behavior, another may prioritize \textit{covariate complexity}, seeking the most plausible explanation for human understanding. Achieving all quality properties simultaneously poses a challenge, as meeting one requirement may conflict with another.

We propose to consolidate each list of axioms in an attempt to synthesize them into a comprehensive and representative compilation. To achieve this, we extracted the axioms expressed from existing surveys, denoted by \remi{the meta survey done by} \cite{le2023benchmarking}. The compiled results are available in Table \ref{table_axioms}. %
By analyzing the axioms listed in Table \ref{table_axioms}, we identified common points that allowed us to categorize them into five overarching axioms: \textit{trustworthiness}, \textit{complexity}, \textit{robustness}, \textit{generalizability}, and \textit{objectivity}:

\begin{table}[!t]
\fontsize{8.5pt}{8.5pt}\selectfont
\caption{\textbf{Common XAI axioms in the literature.}}
\centering
\begin{tabular}{@{}p{4.5cm}p{12cm}@{}}
\toprule
\textbf{Authors} & \textbf{Notions} \\
\midrule
Fel et al. \cite{fel2022good} & Fidelity, Generalizability , Stability, Comprehensibility, Consistency \\
Liao et al. \cite{liao2022connecting} & Faithfulness, Translucence, (Un)Certainty, Interactivity, Stability, Comprehensibility, Completeness, Actionability, Personalization, Coherence, Compactness, Novelty \\
Co-12 \cite{nauta2023anecdotal}& Correctness, Completeness, Consistency, Continuity, Contrastivity, Covariate complexity, Compactness, Composition, Confidence, Context, Coherence, Controllability \\
Arrieta et al. \cite{arrieta2020explainable} & Trustworthiness, Causality, Transferability, Informativeness, Confidence, Fairness, Accessibility, Interactivity, Privacy awareness \\
Ali et al. \cite{ali2023explainable}& Translucency, Portability, Explanatory Power, Algorithmic Complexity, Generalizability, Fidelity, Consistency, Accuracy, Stability, Comprehensibility, Certainty, Interpretability, Representativeness, Explanation using contrastiveness, Specificity, Sociological, Abnormality, Factuality, Fairness, Privacy, Reliability, Causality \\
Akhtar et al. \cite{akhtar2023survey} & Model Fidelity, Localisation, Stability, Conciseness, Sanity preservation, Axiomatic properties \\
Guidotti et al. \cite{guidotti2018survey} & Interpretability, Accuracy, Fidelity \\
Burkart et al. \cite{burkart2021survey} & Trust, Causality, Transferability, Informativeness, Fair and Ethical Decision Making, Accountability, Making Adjustments, Proxy Functionality \\
Doshi-Velez \& Kim et al. \cite{doshi2017towards} & Fairness, Privacy , Reliability, Robustness, Causality, Usability, Trust \\
Zhou et al. \cite{zhou2021evaluating} & Clarity, Broadness, Parsimony, Completeness, Soundness \\
Confalonieri et al \cite{confalonieri2021historical} & Causal, Counterfactual, Social, Selective, Transparent, Semantic, Interactive \\
Markus et al. \cite{markus2021role} & Clarity, Parsimony, Completeness, Soundness \\
Belle et al. \cite{belle2021principles} & Comprehensibility, Fidelity, Accuracy, Scalability, Generality \\
Vilone et al. \cite{vilone2020explainable} & Algorithmic transparency, Actionability, Causality, Completeness, Comprehensibility, Cognitive relief, Correctability, Effectiveness, Efficiency, Explicability, Explicitness, Faithfulness, Intelligibility, Interactivity, Interestingness, Interpretability, Informativeness, Justifiability, Mental Fit, Monotonicity, Persuasiveness, Predictability, Refinement, Reversibility, Robustness, Satisfaction, Scrutability / diagnosis, Security, Selection / simplicity, Sensitivity, Simplification, Soundness, Stability, Transparency, Transferability, Understandability \\
Rojat et al. \cite{rojat2021explainable} & Explainability, Interpretability, Trustworthiness, Interactivity, Stability, Robustness, Reproducibility, Confidence \\
Beaudouin et al. \cite{beaudouin2020flexible} & Accountability, Accuracy, Auditability, Fidelity, Inscrutability, Interpretability, Monotonicity, Robust, Simulatability, Traceability, Transparency, Usability \\
Bennetot et al. \cite{bennetot2022greybox} & Objectivity, Intrinsicality, Validity, Completeness \\
\bottomrule
\end{tabular}
\label{table_axioms}
\end{table}

\begin{itemize}
    \item \textbf{Trustworthiness}: This axiom pertains to the XAI method's ability to accurately reflect the underlying functioning of the model it aims to explain. It encompasses for example notions of accuracy, fidelity, and validity.
    \item \textbf{Robustness}: Robustness refers to the XAI method's resilience against perturbations. It includes aspects \remi{like} consistency and stability.
    \item \textbf{Complexity}: Complexity relates to the XAI method's capacity to provide explanations that are both simple and informative. It involves notions of comprehensibility and confidence.
    \item \textbf{Generalizability}: Generalizability denotes the XAI method's adaptability across a broad range of models. It encompasses aspects \remi{like} transferability and coherence.
    \item \textbf{Objectiveness}: Objectiveness refers to the XAI method's ability to generate explanations that elicit consensus among humans. It includes notions \remi{like} representativeness and certainty.
\end{itemize}

\subsection{Evaluation metrics}

\subsubsection{Overview}

To go further than the simple characterization of methods, researchers have developed metrics to quantify the diverse properties of explanations, resulting in a multitude of measures. This progress in XAI evaluation has prompted the development of numerous libraries aimed at providing researchers with visualizations of metric performances, often presented as arrays of characteristics \cite{hedstrom2023quantus}. These advancements have led to notable progress in the quantitative analysis of XAI methods.

In Table \ref{xai-evaluation}, we present an inventory of the predominant evaluation methods employed in the field of XAI. To compile this inventory, we systematically extracted each evaluation technique used in the methods outlined in Section and \ref{xai-methodsPFMforXAI}. Furthermore, we supplemented this list with the inclusion of widely recognized XAI evaluation methods, as identified through prominent evaluation \remibis{libraries} such as Quantus~\cite{hedstrom2023quantus} and Xplique~\cite{fel2022xplique}. 

\begin{table}[htbp]
\fontsize{8.5pt}{8.5pt}\selectfont
\caption{\textbf{Evaluation methods for XAI.} \textbf{Axioms evaluated} refers to the axioms that the method fulfills according to our desirata and \textbf{Explanation modality} refers to the potential restrictions on the explanation format that the method requests. \textbf{Need GT} refers to the necessity to ground truth explanations to apply the metric.}
\label{xai-evaluation}
\centering
\begin{tabular}{llll}
\toprule
\textbf{Name method} & \textbf{Axioms evaluated} & \textbf{Explanation modality} & \textbf{Need GT} \\
\midrule
Avg-Sensitivity \cite{yeh2019fidelity} & Robustness & All & No \\
BLEU \cite{papineni2002bleu} & Objectiveness & Text & Yes \\
CIDEr \cite{vedantam2015cider} & Objectiveness & Text & Yes \\
CLIP Score \cite{radford2021learning} & Objectiveness & Image, Text & Yes \\ 
Completeness \cite{sundararajan2017axiomatic} & Trustworthiness & All & No \\
Complexity \cite{bhatt2020evaluating} & Complexity & All & No \\
Consistency \cite{dasgupta2022framework} & Robustness & All & No \\
Continuity \cite{montavon2018methods} & Robustness & All & No \\
Deletion \cite{petsiuk2018rise} & Trustworthiness & All & No \\
DSG \cite{chodavidsonian} & Objectiveness & Image, Text & Yes \\
Effective Complexity \cite{nguyen2020quantitative} & Complexity & All & No \\
Efficient MPRT \cite{hedstrom2024sanity} & Generalization & All & No \\
Faithfulness Correlation \cite{bhatt2020evaluating} & Trustworthiness & All & No \\
Faithfulness Estimate \cite{alvarez2018towards} & Trustworthiness & All & No \\
FID \cite{heusel2017gans} & Trustworthiness & Counterfactual Image & No \\
Infidelity \cite{yeh2019fidelity} & Trustworthiness & All & No \\
Input Invariance \cite{kindermans2019reliability} & Trustworthiness & All & No \\
Insertion \cite{petsiuk2018rise} & Trustworthiness & All & No \\
IROF \cite{rieger2020irof} & Trustworthiness & All & No \\
Jaccard similarity \cite{panousis2023hierarchical} & Objectiveness & Concept & Yes \\
Local Lipschitz Estimate \cite{alvarez2018towards} & Robustness & All & No \\ 
LPIPS \cite{zhang2018unreasonable} & Trustworthiness & Counterfactual Image & No \\
Max-Sensitivity \cite{yeh2019fidelity} & Robustness & All & No \\
MeGe \cite{fel2022good} & Generalization & All & No \\
METEOR \cite{banerjee2005meteor} & Objectiveness & Text & Yes \\
Monotonicity Metric \cite{arya2019one} & Robustness & All & No \\
MPRT \cite{adebayo2018sanity} & Generalization & All & No \\
Non-Sensitivity \cite{nguyen2020quantitative} & Trustworthiness & All & No \\
PASTA-metric \cite{kazmierczak2024benchmarking} & Objectiveness & Image, Concept & No \\
Pixel Flipping \cite{bach2015pixel} & Trustworthiness & All & No \\
Random Logit Test \cite{sixt2020explanations} & Generalization & All & No \\
Reasoning metric \cite{nie2023reason2drive} & Trustworthiness & Text & Yes \\
Reasoning performance \cite{chen2023measuring} & Trustworthiness & Text & Yes \\
Recognition accuracy \cite{wan2024interpretable} & Objectiveness & Concept & Yes \\
ReCo \cite{fel2022good} & Robustness & All & No \\
Region Perturbation \cite{samek2016evaluating} & Trustworthiness & All & No \\
RIS \cite{agarwal2022rethinking} & Robustness & All & No \\
ROAD \cite{rong2022consistent} & Trustworthiness & All & No \\
ROUGE \cite{lin2004rouge} & Objectiveness & Text & Yes \\
ROS \cite{agarwal2022rethinking} & Robustness & All & No \\
RRS \cite{agarwal2022rethinking} & Robustness & All & No \\
Selectivity \cite{montavon2018methods} & Trustworthiness & All & No \\
Sensitivity \cite{ancona2017towards} & Trustworthiness & All & No \\
SeeTRUE \cite{yarom2024you} & Objectiveness & Image, Text & Yes \\
Smooth MPRT \cite{hedstrom2024sanity} & Generalization & All & No \\
Sparseness \cite{chalasani2020concise} & Complexity & All & No \\
SPICE \cite{anderson2016spice} & Objectiveness & Text & Yes \\
Sufficiency \cite{dasgupta2022framework} & Trustworthiness & All & No \\
TIFA \cite{hu2023tifa} & Objectiveness & Image, Text & Yes \\
Top K concepts \cite{kalibhat2023identifying} & Objectiveness & Concept & Yes \\
VIEScore \cite{ku2023viescore} & Objectiveness & Image, Text & Yes \\
VisualGPTScore \cite{lin2023revisiting} & Objectiveness & Image, Text & Yes \\
VPEVAL \cite{cho2024visual} & Objectiveness & Image, Text & Yes \\
\bottomrule
\end{tabular}
\end{table}

In Table \ref{xai-evaluation}, we can notice a significant disparity  between the evaluation of \textit{objectiveness} and the other axioms. While trustworthiness, complexity, robustness, and generalizability can be quantitatively assessed by examining the model's response to perturbed inputs, objectiveness presents a distinct challenge. This axiom, which involves human judgment, requires human-labeled data.

\subsubsection{Uses of evaluation metrics in PFMs\remibis{-based} XAI methods}

As reported by \cite{nauta2023anecdotal}, around $58\%$ of research papers in the field have integrated quantitative evaluation methods into their studies. It is noteworthy that this statistic pertains to XAI techniques predating the emergence of PFMs-based approaches.

To investigate trends in the behavior of PFM-based XAI methods, we conducted a similar analysis to the one by~\cite{nauta2023anecdotal}, by assessing the prevalence of quantitative evaluation in recent studies. Our findings reveal that only $36\%$ of the proposed methods include quantitative results.

Significantly, a notable discrepancy exists between different families of explanation methods. For example, while explanations based on text generation commonly employ measures rooted in text alignment metrics, CBMs encounter challenges in extending beyond qualitative explanations.
This observed variance could be explained by divergent perspectives on explainability, often interpreted as the proximity of the decision-making process to human cognition. 
\remi{For instance, some methods, such as in Section \ref{CoT_PFM_for_XAI}, self-identify as explainable due to their incorporation of Chain-of-Throught reasoning, mirroring aspects of human decision-making processes. This aspect of explainability is more difficult to measure compared to feature activation maps,  it is then harder to quantify the quality of the produced explanations.}

Another notable challenge in evaluating PFMs-based XAI methods stems from the diversity of explanation types they generate. While traditional XAI techniques typically produce numerical explanations, PFMs-based approaches can encompass various modalities, including textual and visual outputs. Consequently, specific evaluation metrics tailored to each modality, such as \cite{lin2004rouge,papineni2002bleu} for text-based explanations and \cite{heusel2017gans} for image-based explanations, become essential. \remibis{Additionally, metrics designed to evaluate explanations that span multiple modalities are beginning to emerge.} \remibis{Then, \cite{kazmierczak2024benchmarking} introduces a metric that unifies concept-based and saliency-based evaluation, bridging the gap between these two approaches. Another emerging trend is the development of specialized benchmarks tailored for vision-language tasks, which assess text-image alignment \cite{yarom2024you,cho2024visual,chodavidsonian,lin2023revisiting}. Such datasets are particularly well-suited for evaluating tasks like Visual Question Answering (VQA) and text-to-image generation, ensuring comprehensive testing of multimodal model capabilities.}

\section{Observations} \label{Observations}

\paragraph{Text-image multimodality}
     From a broader viewpoint, image language models are the predominant solutions this survey highlights. This dominance finds its rationale in the maturity of this domain, notably marked by the success of GPT \cite{radford2019language}. Additionally, the intrinsic nature of textual information as a form of explanation contributes to this prevalence. This fusion has facilitated the adaptation and extension of various XAI techniques originally derived from the field of language-based models, such as the CoT methodology.

   \paragraph{Reduction of training requirements} %
     Concerning model-based methods, the integration of PFMs appears to extend the line of methods adapted to dedicated architectures trained on specific datasets. For instance, before PFMs, CBMs were applied to datasets equipped with inherent attributes. However, a paradigm shift is evident in the approach, departing from previous methods that required end-to-end training. The use of PFMs introduces a notable shift from prior approaches in two significant ways:
    \begin{itemize}
        \item Elimination of training on restricted datasets: unlike previous methodologies reliant on constrained datasets, PFMs alleviate the necessity of training on specific, limited datasets. 
        \item Partial model training: leveraging PFMs enables the practice of not training the entire model from scratch. Effective solutions are attainable through the use of frozen or fine-tuned PFMs, removing the need for exhaustive retraining.
    \end{itemize}

\paragraph{Task automation}  %
    Concerning post-hoc methods, there is a notable novelty, stemming from the expanded capabilities enabled by PFMs, such as generating lifelike images from textual inputs. Within the spectrum of post-hoc XAI methods leveraging PFMs, these models consider PFMs as tools capable of executing high-level tasks automatically, such as labeling or generating images.

\section{Challenges} \label{sec:limits_and_challenges}

\remi{In this section, we outline a set of key challenges and limitations associated with PFMs, as identified through a comprehensive analysis of the surveyed literature. Each subsection focuses on a specific aspect of PFMs that either poses a limitation or represents a promising avenue for future research. For each challenge, we highlight \remibis{in Section \ref{chall_current} and Figure \ref{taxonomy_chall}} relevant works that propose solutions or provide insights into addressing these issues. In Section \ref{chall_open}, we address several points that we consider interesting yet untackled in the current literature.}

\subsection{Challenges currently tackled in the state of the art} \label{chall_current}

\begin{figure} 
    \begin{forest}                      
    forked edges,                        
    for tree={grow=0,draw, 
    font=\strut\scriptsize\sffamily\em}, 
        [Challenges currently tackled 
                [Human and PFMs similarity
                    [\scriptsize \cite{yang2023brain}]]
                [Background knowledge
                    [\scriptsize \cite{zhu2023contrastive,opielka2024saliency,martinez2023hypericons,ghiasi2022vision,lin2023revisiting}]]
                [Reasoning capabilities. 
                    [\scriptsize \cite{chen2023measuring,ahrabian2024curious,zhang2024mathverse,chodavidsonian}]]
                [Spurious correlations/bias
                    [\scriptsize \cite{he2024tubench,giledereli2024vision,jiang2024texttt,gao2024vision,zhang2024common,moayeri2024spuriosity}]]
                [Adaptation of feature saliency-based techniques
                    [\scriptsize \cite{chefer2021generic,wang2024visual,park2024explaining,li2023clip,chen2022gscorecam,li2022exploring,arya2024b,dewan2024diffusionpid,bousselham2024legrad}]]]
    \end{forest}
    \caption{Summary of works that tackle issues raised by the use of PFM in XAI.}
    \label{taxonomy_chall}
\end{figure}
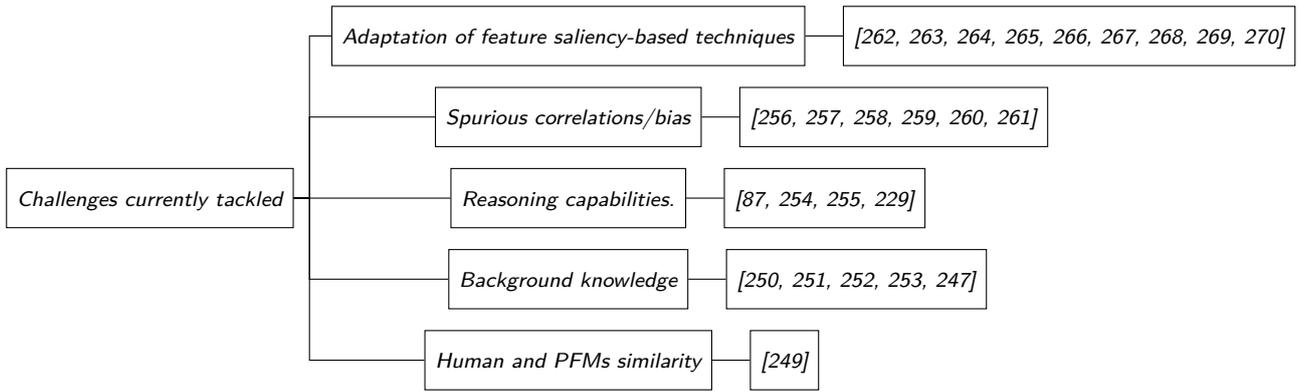

\subsubsection{\remi{Adaptation of actual attribution based techniques}}

\remi{Even today, post-hoc attribution methods, such as GradCAM \cite{selvaraju2017grad}, \remibis{B-cos \cite{bohle2022b}} or SHAP \cite{lundberg2017unified}, remain some of the most widely used techniques for explaining deep neural networks (DNNs) in computer vision. While these methods are recognized for their robustness, they were primarily designed and evaluated on older architectures, such as ResNet50 \cite{he2016deep}, which may introduce a bias favoring these networks \cite{kazmierczak2024benchmarking}.
The increasing prevalence of transformer-based architectures, which form the backbone of many PFMs, necessitates revisiting these attribution techniques to address potential shortcomings. For instance, \cite{li2022exploring,li2023clip} identified a phenomenon called pixel flipping in CLIP models, where attention maps can unintentionally invert, undermining their interpretability. Efforts to adapt saliency-based techniques for CLIP, such as GScoreCAM \cite{chen2022gscorecam} \remibis{or LeGrad \cite{bousselham2024legrad}}, exemplify the need for tailored approaches. \remibis{Concerning B-cos, \cite{arya2024b} proposes a variant of the original method that does not require retraining a model, allowing B-cosification of PFMs.}
Similarly, emerging architectures like diffusion models have prompted new explorations of saliency-based explanations, as demonstrated by \cite{park2024explaining} \remibis{or \cite{dewan2024diffusionpid}}. Another pressing challenge is the multimodal nature of modern PFMs, which process both image and text inputs. Methods must evolve to simultaneously highlight relevant pixels and text tokens used during inference \cite{wang2024visual,chefer2021generic}.}

\subsubsection{\remi{Spurious correlations/bias}} \label{spur_corr}

\remi{A critical aspect of PFMs lies in their ability to base inferences on relevant and meaningful features. For example, in image classification tasks, we expect models to classify images by recognizing features intrinsic to the target object. However, this expectation is often violated, as illustrated by the well-known ``Wolf vs. Husky'' example \cite{lime}, where a model incorrectly focuses on the presence of snow—a spurious feature correlated with images of huskies—rather than on the actual object features. This highlights the prevalence of biases and spurious correlations in model predictions, prompting numerous studies to investigate their extent in PFMs.
For instance, tailored datasets have been used to uncover biases in models like CLIP, revealing systematic issues regarding spurious correlations \cite{moayeri2024spuriosity,zhang2024common}. Similarly, visual question-answering (VQA) PFMs, including BLIP, GPT-4, and LLaVa, exhibit notable shortcomings. Research has identified failures in handling unanswerable questions \cite{he2024tubench}, visual diagrams \cite{giledereli2024vision}, stereotypical biases \cite{jiang2024texttt}, and intent recognition \cite{gao2024vision}.
Understanding these biases is crucial, especially given the extensive use of these models as foundations for methods discussed in Section~\ref{xai-methodsPFMforXAI}. }

\subsubsection{\remi{Reasoning capabilities}} \label{CoT_XAI_for_PFM}

\remi{In addition to identifying biases in VQA PFMs, there is growing concern about their reasoning capabilities. To address this, several benchmarks have been developed to evaluate reasoning in diverse contexts, including daily scenes \cite{chen2023measuring}, puzzles \cite{gao2024vision}, and mathematical problems \cite{zhang2024mathverse}. Results from these studies indicate that VQA PFMs face significant challenges when attempting to solve such tasks in a zero-shot setting. However, the use of chain-of-thought (CoT) prompting has been shown to substantially improve performance, aligning with observations made for certain models discussed in Section~\ref{CoT_PFM_for_XAI}.}

\subsubsection{\remi{Background knowledge}}

\remi{A related concept to spurious correlations and biases is the notion of background knowledge. Introduced by \cite{giledereli2024vision}, this term refers to the global understanding of concepts that PFMs acquire through training on extensive datasets. According to \cite{giledereli2024vision}, background knowledge often leads PFMs to use shortcuts when answering questions, particularly for tasks involving visual diagrams, resulting in erroneous responses. Identifying and analyzing such background knowledge has become an active area of research.
For instance, works like \cite{ghiasi2022vision} and \cite{martinez2023hypericons} investigate optimizing CLIP or stable diffusion neurons to activate specific abstract concepts, employing techniques akin to those discussed in Section~\ref{sec:neur_interpret}. Another approach involves creating tailored datasets designed to decouple saliency and semantics \cite{opielka2024saliency} or exploring prompt preferences in diffusion models to reveal latent tendencies.}

\subsubsection{\remi{Alignment with Human Cognition}}

\remi{Finally, another emerging challenge is understanding the similarities between human brain processes and PFMs. While this topic has garnered interest, research remains scarce due to the high costs and experimental complexities involved. A notable example is the work of \cite{yang2023brain}, which investigates the relationship between neural activations in various PFM backbones and brain MRI responses to identical image stimuli.}

\subsection{\remibis{Open challenges}} \label{chall_open}

\subsubsection{Towards more \remibis{mathematicaly grounded explanations}}
     From a comprehensive perspective, XAI methods based on PFMs prioritize functionality over mathematical groundings. Traditional XAI methods like SHAP rely on mathematical theories to elucidate their operations, yielding explanations in the form of numerical values. In contrast, PFMs-based XAI methods tend to offer comprehensive systems, such as textual captions, as explanations. While this approach to designing XAI methods promotes understandability and facilitates explanations for highly intricate models, it inevitably sparks questions regarding trustworthiness and fidelity.

\subsubsection{\remibis{Towards incorporation of new modalities}}

Existing methods are closely linked to the capabilities of current PFMs, yet the advent of novel approaches has expanded the scope of potential applications. A notable example is SAM \cite{zou2024segment}, which enables zero-shot image segmentation.

In addition, just as language reinforces the development of image-centric XAI, the recent incorporation of new modalities like depth or sound \cite{girdhar2023imagebind} holds the promise of advancing explainability in AI systems. However, a notable limit in this domain is the challenge of generating datasets that match the scale and richness of existing image and vision datasets.

\subsubsection{Quantitative evaluation}
This trend can also be observed on the side of evaluation, where studies on PFMs-based methods tend to include fewer measurements beyond qualitative examples. Notably, objectiveness is often measured, but other axioms are much less evaluated. \remibis{In a global manner, PFMs-based XAI methods struggle to be evaluated quantitatively, notably due to the difficulty of conciliating multimodality.}

\subsubsection{Detect spurious explanations} 

\remibis{Given the inherent propensity of PFMs to exhibit biases (see Section \ref{spur_corr}), it is reasonable to hypothesize that PFM-based XAI methods may be susceptible to producing spurious explanations—those that rely not on genuine correlations within a given input, but rather on implicit associations learned during the pretraining of the foundation model. For instance, can we confidently assert that the concept scores generated by a CLIP-based CBM correspond to actual patterns present in an image? Consequently, an open challenge in this domain lies in the identification and mitigation of such potential spurious explanations.}

\section{Conclusions} \label{Conlusion}

The growing interest in integrating explainability into large models is undoubtedly commendable, particularly \remi{considering that frequently used methods remain} entirely opaque  \cite{brown2020language}. However, it is crucial to acknowledge an inherent tension regarding the role of PFMs in XAI methods. While PFMs are lauded for their capacity to accomplish high-level tasks like multimodal integration, their utilization demands caution due to the inherent opacity they introduce in the model.

Current efforts to explain PFMs have resulted in substantial progress. However, pursuing a comprehensive deep-learning theory that renders PFMs transparent appears to be a distant goal. Then, an alternative approach to gain more control over PFMs-based XAI methods involves focusing not on understanding the inner workings of PFMs themselves but on modeling their outputs, namely the latent spaces they produce. Previous research \cite{radford2021learning} underscores the remarkable effectiveness of PFMs in transforming high-entropy data, such as image datasets, into significantly lower-entropy embeddings. Moreover, these latent spaces appear amenable to modeling through simple distributions \cite{kazmierczakclip}. Advancements in modeling the latent space hold considerable promise for the framework \remibis{of inherently explainable models}. By gaining insights into the distribution process within the transparent segment of the model, it becomes feasible to conceptualize our entire algorithm as the combination of a opaque model and a mathematically explainable model.

Another notable observation is the tendency to justify explainability through the notion of human-like reasoning, as described in Section \ref{CoT_PFM_for_XAI}. However, it is essential to exercise caution with such arguments, as the parallel between DNNs and the human brain remains largely unproven. Particular caution must also be  kept in the use of PFMs for computing high-level tasks in post-hoc methods. While PFMs offer generalization properties that render them suitable for tasks like labeling, users must remain conscious  of the absence of guarantees regarding their performance across all use cases.

\section{Acknowledgements}

This research has been financed by a Hi!Paris and ANR IA PhD grant.

\bibliographystyle{elsarticle-num-names} 
\bibliography{elsarticle-template-num-names.bib}

\end{document}